\documentclass[10pt,letterpaper,twocolumn]{article}

\usepackage[textwidth=6.5in,textheight=9in,centering,headheight=14pt,headsep=18pt]{geometry}
\usepackage{tgpagella}
\usepackage{mathpazo}
\usepackage{helvet}
\usepackage{courier}
\usepackage[hyphens]{url}
\usepackage{graphicx}
\usepackage{natbib}
\usepackage{caption}
\usepackage{placeins}
\usepackage{booktabs}
\usepackage{amsfonts}
\usepackage{nicefrac}
\usepackage{microtype}
\usepackage{array}
\usepackage{pifont}
\usepackage{makecell}
\usepackage{multirow}
\usepackage{amsmath}
\usepackage{xspace}
\usepackage{fontawesome}
\usepackage[table]{xcolor}
\usepackage[most]{tcolorbox}
\usepackage{fancyhdr}
\usepackage[colorlinks=true,linkcolor=blue,citecolor=blue,urlcolor=blue]{hyperref}

\definecolor{sharedbudget}{HTML}{008080}
\newcommand{\sharedbudgets}{\textcolor{sharedbudget}{Shared Budgets}}

\definecolor{darkgreen}{rgb}{0.0,0.55,0.0}
\definecolor{darkred}{rgb}{0.70,0.0,0.0}
\definecolor{tsinghuapurple}{RGB}{102,8,116}
\newcommand{\cmark}{\textcolor{darkgreen}{\ding{51}}}
\newcommand{\xmark}{\textcolor{darkred}{\ding{55}}}
\newcommand{\bench}{\textsc{$R^3$-Bench}\xspace}
\newcommand{\best}{\cellcolor[HTML]{B8E7E1}}
\newcommand{\second}{\cellcolor[HTML]{D4FAFC}}
\newcommand{\codeurl}{https://github.com/NineAbyss/R-3-Bench}
\newcommand{\dataseturl}{https://huggingface.co/datasets/R-3-Bench/R-3-Bench}

\newtcolorbox{promptbox}[3][Prompt]{
  colback=black!5!white,
  arc=5pt,
  boxrule=0.5pt,
  fonttitle=\bfseries,
  title=#1,
  before upper={\small},
  fontupper=\fontfamily{ppl}\selectfont,
  colframe=#2,
  label=#3,
  breakable
}

\makeatletter
\newcommand{\appendixcontents}{%
  \begingroup
  \small
  \renewcommand*\numberline[1]{\makebox[1.5em][l]{##1}}%
  \renewcommand*\l@section[2]{%
    \noindent\hangindent=1.5em\hangafter=1
    ##1\nobreak\dotfill\nobreak\makebox[2em][r]{##2}\par}
  \@starttoc{apc}%
  \endgroup}
\newcommand{\enableappendixcontents}{%
  \let\arxivoriginaladdcontentsline\addcontentsline
  \renewcommand{\addcontentsline}[3]{%
    \arxivoriginaladdcontentsline{##1}{##2}{##3}%
    \def\appendixentrytype{##2}%
    \def\appendixsectiontype{section}%
    \ifx\appendixentrytype\appendixsectiontype
      \addtocontents{apc}{%
        \protect\contentsline{##2}{##3}{\thepage}{\@currentHref}}%
    \fi}}
\makeatother

\hypersetup{
  pdftitle={R3-Bench: LLMs Struggle with Resource-Rational Reasoning under Shared Budgets},
  pdfauthor={Peisong Wang, Zhiwei Ma, Bowen Liu, Feixue Liu, Aochuan Chen, Chenyi Zi, Hongchuan Zeng, Yuhan Li, Jia Li}
}
\title{\raggedright\Large\bfseries
  \raisebox{-0.25\height}{%
    \includegraphics[height=1.8em]{figs/profileR3.pdf}}%
  \hspace{0.35em}%
  {\itshape $R^3$-Bench:} LLMs Struggle with Resource-Rational Reasoning\\
  under \sharedbudgets\par}
\author{
  \normalfont\normalsize\bfseries
  Peisong Wang\textsuperscript{1}\thanks{Equal contribution.}\quad
  Zhiwei Ma\textsuperscript{1}\footnotemark[1]\quad
  Bowen Liu\textsuperscript{1}\quad
  Feixue Liu\textsuperscript{2}\quad
  Aochuan Chen\textsuperscript{1}\quad\\
  \normalfont\normalsize\bfseries
  Chenyi Zi\textsuperscript{1}\quad
  Hongchuan Zeng\textsuperscript{3}\quad
  Yuhan Li\textsuperscript{1}\quad
  Jia Li\textsuperscript{1}\thanks{%
  Correspondence to: Jia Li
  \textless{}jialee@hkust-gz.edu.cn\textgreater{}.}\\[5pt]
  \normalfont\normalsize
  \textsuperscript{1}The Hong Kong University of Science and Technology (Guangzhou)\\
  \normalfont\normalsize
  \textsuperscript{2}The University of Hong Kong\qquad
  \textsuperscript{3}Hunyuan Team, Tencent\\
  \normalfont\normalsize
  \texttt{\{ps.wang,zma191\}@connect.hkust-gz.edu.cn}\\[3pt]
  \normalfont\normalsize
  \href{\codeurl}{\faGithub\ Code}
  \hspace{2em}
  \href{\dataseturl}{%
    \raisebox{-0.15\height}{\includegraphics[height=1.1em]{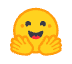}}\ Dataset}
}
\date{}

\begin{document}

\maketitle
\thispagestyle{fancy}

\begin{abstract}
In cognitive science, resource rationality asks how an agent should allocate
limited computation to maximize expected value. Most reasoning and agent
benchmarks use independent per-task budgets; existing shared-budget studies
do not calibrate suite performance against the same model's demonstrated
single-problem competence. We introduce \bench{}, which evaluates six-problem
suites under shared budgets across mathematics, competitive programming, and
abstract reasoning in tool-free and agentic settings. Matched single-problem
response curves define an offline empirical oracle over observed successes.
Across 72 main-table cells for six models, the oracle mean matches or exceeds
the contest mean in all cells and is strictly higher in 71. Under moderate
tool-free pressure, equal-allocation replay also exceeds contest performance
for three of six models. Trajectory diagnostics reveal limited strategy
updating and pressure-dependent failure patterns. In a three-model diagnostic
under strong agentic pressure, at least one fixed scheduler exceeds the
contest mean in six of nine cells, but no policy dominates across domains.
These results expose a persistent gap between demonstrated competence and
shared-budget realization.
\end{abstract}

\section{Introduction}

Reasoning has become a central capability of modern large language models (LLMs). 
Chain-of-thought prompting and test-time scaling show that additional
inference-time computation can improve LLM
reasoning~\citep{wei2022chain,snell2024scaling}. These gains are not free:
reasoning consumes tokens, wall-clock time, tool calls, and verifier
interactions. Under resource constraints, an agent must decide not only how to
solve a problem, but also whether further computation is worthwhile---whether
to think longer, call a tool, run a test, verify a candidate, switch tasks, or
stop.
\begin{figure}[!t]
    \centering
    \includegraphics[width=0.95\columnwidth]{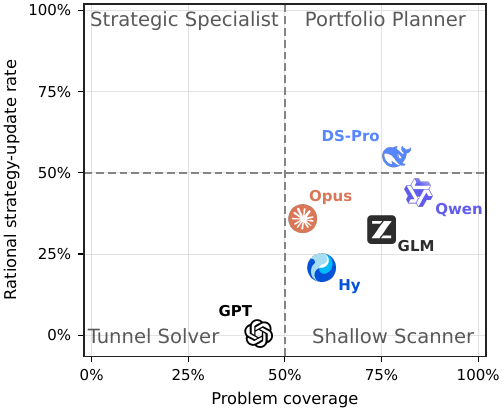}
    \caption{Problem coverage and resource-rational strategy-update rate for the six flagship models, aggregated across tool-free and agentic settings, both budget pressures, and all available domains. Coverage is the share of problem slots with visible, substantive problem-specific work; resource-rational updates are strategy changes that respond to new evidence and the remaining shared budget. Dashed lines mark descriptive 50\% references on each metric. Appendix~\ref{app:behavioral-coordinate} provides the complete definitions.}
    \label{fig:coo}
\end{figure}
This is a resource-rationality problem. Rational analysis models cognition as
balancing expected gains against computational
costs~\citep{anderson1991human}. Resource-rational analysis makes this
constraint explicit by viewing cognition as the optimal use of
\emph{limited computational 
resources}~\citep{lieder2020resource}. 

For LLM systems, this resource-allocation problem often extends beyond a single task.
Real agent workflows increasingly involve several tasks or agents operating concurrently. More than 10\% of Codex users manage at least three concurrent
agents in a given week~\citep{johnston2026shift}. An analysis of roughly
400,000 Claude Code sessions further shows that coding-agent workflows
routinely involve building, debugging, testing, and operating
software~\citep{hitzig2026agentic}. Such workloads draw on finite API quotas, inference capacity, and external endpoints; resource contention among concurrent tasks can therefore lead to wasted computation~\citep{agyemang2026hivemind}.
Once several tasks share a finite inference budget, computation spent on one task reduces the resource available to the others, creating an opportunity cost and making cross-task allocation part of the reasoning problem itself.

Existing benchmarks make this ability hard to observe.
Reasoning benchmarks evaluate one problem at a time~\citep{hendrycksmath2021,rein2023gpqa,tang2025grapharena,zheng2025livecodebench}, and agent benchmarks
typically budget each task in isolation~\citep{liu2024agentbench,jimenez2024swe,mialon2024gaia,zhang2025sentient}.
In either setting, the cross-problem allocation question never arises: a model can spend its full allowance on every problem without considering the remaining tasks.
We therefore ask whether an LLM can realize its demonstrated single-problem
competence when computation has an opportunity cost across problems?

To answer this question, we introduce \bench{}, a benchmark for
\textbf{\underline{R}}esource-\textbf{\underline{R}}ational
\textbf{\underline{R}}easoning in LLM systems. For each of three domains---
Olympiad-style mathematics, competitive programming, and abstract
reasoning---\bench{} constructs 50 six-problem contests containing problems with different resource demands. Each contest places all six problems under one shared
budget and is evaluated in both tool-free and agentic settings. To separate
problem-solving competence from its realization under a shared budget, we
use matched single-problem runs to construct an equal-allocation replay and
a same-model response-curve oracle. The oracle selects demonstrated
single-problem successes subject to the contest's total budget; it is an
offline empirical diagnostic rather than an executable policy.

The results reveal a persistent gap between demonstrated competence and
shared-budget performance. Across the 72 flagship
model--setting--pressure--domain cells reported in the main table, the oracle
matches or exceeds the contest score in every cell and is strictly higher in
71. Under moderate tool-free budget pressure, equal allocation also
outperforms the contest policy for three of six models. Neither a larger
budget nor access to tools uniformly closes the gap. Figure~\ref{fig:coo} provides a complementary behavioral view: five of six models cover more than half of the suite, whereas only DS-Pro exceeds
the descriptive 50\% reference for resource-rational strategy updates. 
Broad coverage therefore does not imply resource-rational behavior. 

Trajectory analysis further shows that failure patterns
depend on budget pressure: among problems selected by the oracle but missed in the actual contest, failures under strong pressure are most often due to exhausting the shared budget on other problems,
whereas under moderate pressure they more often come from stopping after
partial progress. Both categories concern problems the model has already solved in isolation, so both point to allocation---which problems
receive budget, and how much they receive---rather than to a capability limit. The gap is also not a restatement of general capability: models
within two points on a composite of over fifty benchmarks differ by up to
\(60.6\) percentage points in Gap Ratio (Section~\ref{sec:main_results}).
We further probe lightweight online scheduling, finding higher mean performance in six of nine model--domain cells but no policy that transfers uniformly across domains (Section~\ref{sec:recovering-allocation-gap}).
Our contributions are:
\begin{itemize}
    \item We formulate resource-rational reasoning as shared-budget
    task-suite evaluation and introduce \bench{}, spanning three domains and both tool-free and agentic settings.

    \item We use matched single-problem response curves and equal-allocation
    replay to reveal persistent gaps between demonstrated competence and
    shared-budget performance across models and budget pressures, and show that
    these gaps are left unexplained by a general-capability composite over
    fifty benchmarks.

    \item We diagnose pressure-dependent failure patterns and limited online
adaptation, and find domain-dependent effects from fixed online schedulers.
\end{itemize}
\section{Related Work}
\newcolumntype{L}[1]{>{\raggedright\arraybackslash}m{#1}}
\newcolumntype{C}[1]{>{\centering\arraybackslash}m{#1}}

\begin{table}[!t]
    \centering
    \small
    \setlength{\tabcolsep}{3.5pt}
    \renewcommand{\arraystretch}{1.12}

    \caption{\textbf{Comparison with budget-aware and resource-allocation-oriented LLM evaluations.}
    A checkmark indicates explicit budget-aware evaluation in the corresponding
    setting. ``Tool-free'' denotes direct model reasoning without external tool
    interaction. ``Shared budget'' indicates that multiple tasks compete for
    one common budget, rather than receiving independent per-task budgets or
    allocating resources only within a single task.}
    \label{tab:related-budget-benchmarks}

    \resizebox{\columnwidth}{!}{%
    \begin{tabular}{
        L{4.0cm}
        C{1.25cm}
        C{1.05cm}
        C{0.85cm}
        C{0.85cm}
        C{1.75cm}
        C{1.25cm}
    }

        \toprule[1.3pt]

        \multicolumn{1}{C{4.0cm}}{%
            \textbf{Benchmark / Study}
        }
        &
        \multicolumn{5}{c}{%
            \begin{tabular}{
                @{}
                C{1.25cm}
                C{1.05cm}
                C{0.85cm}
                C{0.85cm}
                C{1.75cm}
            }
                \multicolumn{2}{c}{%
                    \makebox[0pt][c]{\textbf{Evaluation Setting}}
                }
                &
                \multicolumn{3}{c}{%
                    \makebox[0pt][c]{\textbf{Domain}}
                }
                \\[-0.25ex]

                \cmidrule(lr){1-2}
                \cmidrule(lr){3-5}

                \makecell[c]{\textbf{Tool-}\\\textbf{Free}}
                &
                \textbf{Agentic}
                &
                \textbf{Math}
                &
                \textbf{Code}
                &
                \makecell[c]{\textbf{Abstract}\\\textbf{Reasoning}}
            \end{tabular}
        }
        &
        \makecell[c]{%
            \textbf{Shared}\\
            \textbf{Budget}
        }
        \\

        \midrule[0.55pt]

        \citet{wang2024reasoning}
        &
        \cmark
        &
        \xmark
        &
        \cmark
        &
        \xmark
        &
        \cmark
        &
        \xmark
        \\

        TALE~\citep{han2025token}
        &
        \cmark
        &
        \xmark
        &
        \cmark
        &
        \xmark
        &
        \xmark
        &
        \xmark
        \\

        SelfBudgeter~\citep{li2025selfbudgeter}
        &
        \cmark
        &
        \xmark
        &
        \cmark
        &
        \xmark
        &
        \xmark
        &
        \xmark
        \\

        BATS~\citep{liu2025budget}
        &
        \xmark
        &
        \cmark
        &
        \xmark
        &
        \xmark
        &
        \xmark
        &
        \xmark
        \\

        General AgentBench~\citep{li2026benchmark}
        &
        \xmark
        &
        \cmark
        &
        \xmark
        &
        \cmark
        &
        \cmark
        &
        \xmark
        \\

        ZEBRA~\citep{hamri2026zebra}
        &
        \xmark
        &
        \cmark
        &
        \xmark
        &
        \cmark
        &
        \xmark
        &
        \xmark
        \\

        USACOArena~\citep{zhou2026creditbudgeted}
        &
        \xmark
        &
        \cmark
        &
        \xmark
        &
        \cmark
        &
        \xmark
        &
        \cmark
        \\

        CLEAR~\citep{wan2026shadow}
        &
        \cmark
        &
        \xmark
        &
        \cmark
        &
        \cmark
        &
        \xmark
        &
        \cmark
        \\
        \midrule[0.75pt]

        \rowcolor{black!5}
        \bench{} (\textbf{ours})
        &
        \cmark
        &
        \cmark
        &
        \cmark
        &
        \cmark
        &
        \multicolumn{1}{
            >{\columncolor{black!5}[\tabcolsep][2\tabcolsep]}
            C{1.75cm}
        }{\cmark}
        &
        \cmark
        \\

        \bottomrule[1.3pt]

    \end{tabular}%
    }
\end{table}

\paragraph{Per-task reasoning and agent evaluation.}
Reasoning benchmarks span domains such as mathematics, science question
answering, graph problems, and competitive programming~\citep{hendrycksmath2021,
rein2023gpqa,tang2025grapharena,zheng2025livecodebench}, and agent benchmarks
add tool use, web interaction, and long-horizon software execution~\citep{liu2024agentbench,
mialon2024gaia,jimenez2024swe,zhang2025sentient}.  Both nevertheless evaluate
and budget tasks independently, measuring per-task competence rather than how
computation is distributed across a task suite.

\paragraph{Budget-aware reasoning and resource allocation.}
Budget-aware methods adapt sampling, token, or tool-use budgets within an individual problem~\citep{wang2024reasoning,han2025token,li2025selfbudgeter,
liu2025budget,lou2025adacot}, while orchestration methods allocate resources across models or pipeline stages~\citep{hamri2026zebra}. In neither case do multiple problems compete for one shared budget. The closest prior works are USACOArena~\citep{zhou2026creditbudgeted} and
CLEAR~\citep{wan2026shadow}. USACOArena studies agent behavior in a credit-budgeted coding arena, while CLEAR optimizes token allocation across queries using an external utility model and a shadow-price policy. However, neither of them measures
allocation against per-task reasoning ability. 
Table~\ref{tab:related-budget-benchmarks} compares representative
budget-aware and resource-allocation-oriented LLM evaluations
across evaluation settings, domains, and whether multiple tasks
compete for one shared budget. Checkmarks indicate protocol
coverage rather than methodological superiority. 
\bench{} is built around that comparison: how
much of a model's competence on isolated problems is realized by its own shared-budget behavior? It combines shared-budget suites with matched single-problem response curves, equal-allocation replay, and an offline empirical oracle across three domains and both tool-free and agentic settings.
\section{Methodology: \bench{}}
\label{sec:benchmark-construction}
\begin{figure*}[!t]
    \centering
    \includegraphics[width=\textwidth]{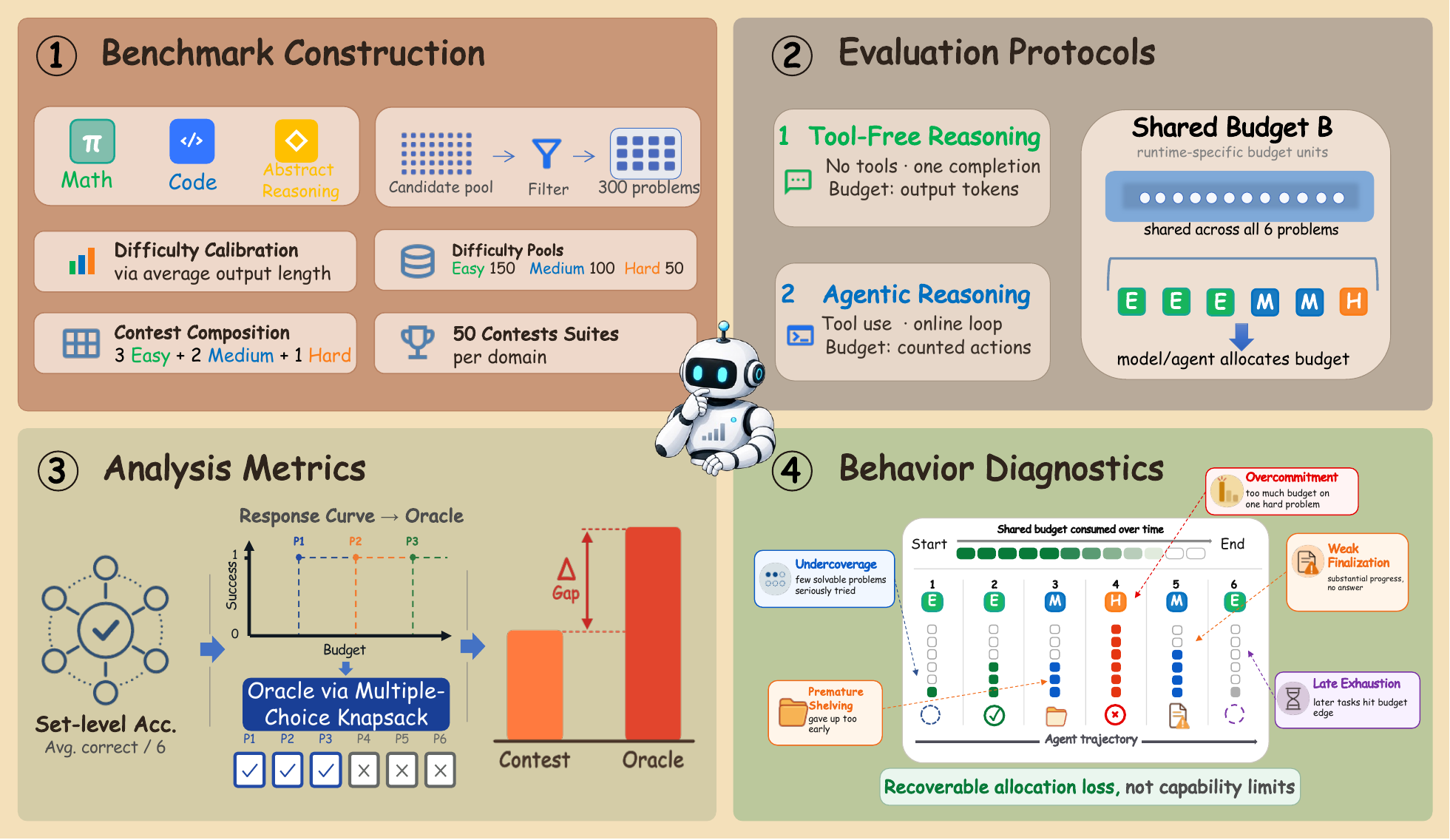}
    \caption{\textbf{Overview of \bench{}.}
    \textcircled{\scriptsize 1} We construct 50 six-problem contests per domain
    from frozen 300-problem pools stratified by the average output length of three
    reference models. Each contest contains three Easy, two Medium, and one Hard
    problem.
    \textcircled{\scriptsize 2} The same contests are evaluated in
    tool-free reasoning and agentic settings under a shared budget, measured in output
    tokens or counted actions.
    \textcircled{\scriptsize 3} Per-problem response curves define an offline
    multiple-choice knapsack oracle that grants each problem one budget level.
    The contest--oracle gap measures demonstrated competence lost under the
    shared budget.
    \textcircled{\scriptsize 4} Trajectory diagnostics characterize how agents
    observe feedback, revise strategies, and allocate the shared budget.}
    \label{fig:benchmark_overview}
\end{figure*}
We construct \bench{} as a suite-level benchmark for evaluating
resource-rational behavior under a shared budget. Each episode is a six-problem \emph{contest}, inspired by multi-problem
programming competitions ICPC and
IOI~\citep{icpc2026rules,ioi2002format}, where participants allocate a fixed time budget across problems of varying difficulty.
\bench{} transfers this time-allocation structure to computation allocation: all six problems share a single resource budget, so
spending resources on one problem reduces what remains for the others. Unlike standard per-problem evaluation, this creates explicit
opportunity costs across problems and requires the model to decide which problems to attempt, how much computation to invest in each,
and when to continue, switch, or stop.

As illustrated in  Figure~\ref{fig:benchmark_overview}, \bench{} contains two evaluation settings. The \emph{tool-free reasoning} setting evaluates models as free-form generators under output-token budgets. The \emph{agentic} setting evaluates models as tool-using agents under action budgets in an interactive shell environment. Both settings share the same problem pools, difficulty tiers, contest construction, answer parser, and grading protocol; they differ only in the runtime and the unit of budget.

\subsection{Task-Suite Construction}
\label{subsec:shared-construction}

\paragraph{Domains and problem pools.}
For each domain, we draw 300 problems: mathematics from
Omni-MATH~\citep{gao2025omni} and MathNet~\citep{alshammari2026mathnet},
competitive programming from LiveCodeBench
Pro~\citep{zheng2025livecodebench}, and abstract reasoning from Reasoning
Gym~\citep{stojanovski2025reasoning}.
The same 300 problems are used for both per-task and
contest-level evaluation.
Appendix~\S\ref{app: data_source} details the sources.

\paragraph{Length-based difficulty stratification.}
We define difficulty by output length rather than post-hoc accuracy.
In a resource-rational setting, difficulty is operationally tied to
resource demand: a problem is more difficult if unbudgeted reference
models naturally spend more output tokens on it.

To reduce dependence on any one model, we average standalone output lengths
from three reference models: DeepSeek V4
Pro~\citep{deepseekai2026deepseekv4},
GLM-5.2~\citep{zai2026glm52}, and
GPT-5.5~\citep{openai2026gpt55}. Within each pool, we label the shortest 150
problems Easy, the next 100 Medium, and the remaining 50 Hard. These fixed
length-based tiers are shared across evaluated models, used only for
construction and analysis, and hidden from the prompts.

\paragraph{Contest construction.}
After stratification, we construct 50 contests per domain, each containing three Easy, two Medium, and one Hard problem. We keep this difficulty mixture fixed across contests to prevent variation in suite composition from confounding comparisons across models and budgets. The heterogeneous
mixture creates a controlled coverage--depth trade-off: a model can spread its budget across several lower-demand problems or invest more deeply in harder ones. Presentation orders are randomized independently of difficulty.

\paragraph{Task formats.}
Both tool-free reasoning and agentic evaluation use the same two
task formats. Difficulty tiers appear in neither prompt; they are used only to
balance contests and analyze performance.

In \emph{per-task} evaluation, each problem is presented independently,
measuring standalone problem-solving ability without cross-problem resource
allocation.

In \emph{contest} evaluation, the model receives all six problems in one
episode and decides which to attempt and how much shared budget to allocate
to each. 

\paragraph{Model-specific budget calibration.}
\label{para:budget-calibration}
Models differ substantially in their natural resource-use patterns: some
produce longer reasoning traces, while others require more tool or verifier actions to make progress. A single absolute budget would therefore impose unequal pressure across models and confound allocation quality with a model's native output pattern. Our goal is not to equalize absolute resource limits, but to evaluate each model under the same relative budget pressure. We
therefore calibrate the budget against each model's own unbudgeted resource consumption.

Let \(R_{m,d,c}^{\infty}\) denote the resources used by model \(m\) on contest
\(c\) in domain \(d\) during an unbudgeted baseline run. Resources are measured as output tokens in the tool-free reasoning setting and counted actions in the agentic setting. Because runs may still hit a runtime safety cap, we compute the baseline using only contests that complete successfully:
\[
R_{m,d}^{\infty}
=
\frac{1}{|\mathcal{C}_{m,d}^{\mathrm{valid}}|}
\sum_{c \in \mathcal{C}_{m,d}^{\mathrm{valid}}}
R_{m,d,c}^{\infty},
\]
where \(\mathcal{C}_{m,d}^{\mathrm{valid}}\) is the set of completed contests.
For pressure level \(\rho \in \{0.2,0.8\}\), we set \(R_{m,d}^{\rho}=\rho R_{m,d}^{\infty}\), rounding when the resource unit is discrete. Thus, \(\rho=0.2\) restricts a model to 20\% of its natural resource use and represents strong pressure, whereas \(\rho=0.8\) represents moderate pressure.

\subsection{Tool-free reasoning Setting}
\label{subsec:pure-nl-reasoning}

The tool-free reasoning setting evaluates models without tools, shell access, code execution, or interactive feedback. The model receives a prompt and produces a single free-form completion. The budget is measured in output tokens.

\paragraph{Budgeted contest evaluation.}
Here \(R_{m,d}^{\rho}\) is a token budget \(B_{m,d}^{\rho}\), passed to the API as \texttt{max\_tokens} and shown in the prompt as one shared allowance.

\subsection{Agentic Setting}
\label{subsec:agentic-benchmark}
The agentic setting evaluates tool-using models in a Terminus-2 harness
implemented with Harbor~\citep{Harbor_Framework}. Agents can execute commands and code, inspect intermediate results, and submit final answers
through designated artifacts. Resource use is measured in counted tool actions. Official correctness feedback is unavailable during the run; correctness is determined only after the episode from the submitted artifacts.

\paragraph{Problem bookkeeping.}
A global contest-level action budget makes per-problem attribution ambiguous. We therefore add two bookkeeping commands:
\texttt{focus\_problem <id>} and \texttt{shelve\_problem} to mark
the active problem. The agent is instructed to call \texttt{focus\_problem <id>} before acting on problem $<$id$>$ and \texttt{shelve\_problem} before switching problems. These commands
are logged separately as \texttt{free bookkeeping steps} and do not consume the shared action budget. They enable per-problem attribution of counted actions, allowing us to analyze how agents spend the shared action budget across Easy, Medium, and Hard problems.

\paragraph{Budgeted contest evaluation.}
In the agentic setting, the action budget is enforced over parsed executable actions rather
than model turns, with each executed counted action consuming one budget unit. Only tool actions that perform problem-solving computation count toward
$R^\rho_{m,d}$; routine file operations remain free. The agent is given the total budget and, through a runtime budget reminder, its
used and remaining counts, so it acts against a visible balance (Appendix~\ref{app:prompts}). Once the budget is exhausted, the runtime blocks further counted actions but permits bookkeeping
and finalization. Correctness is determined from the submitted artifacts. Appendix~\ref{app:white_list} provides the complete action-counting rules.

\subsection{Metrics}
\label{subsec:metrics}

\paragraph{Task success.}
Both settings use the same domain-specific graders and the same binary full-credit criterion. A problem receives one point if its
final answer is judged correct and zero otherwise. Our primary outcome is the \emph{contest score}: the average number of correct answers per six-problem contest. When analyzing performance by difficulty tier or presented position, we additionally report
problem-level accuracy.

\paragraph{Allocation quality.}
We run each problem in isolation on a fixed grid of budget levels, with five independent runs per level, so that its response curve records the empirical success rate at each level. The \emph{equal-allocation replay} asks whether each problem has an observed successful attempt whose realized resource cost fits within one sixth of the contest budget, and serves as a fixed uniform-allocation baseline. The \emph{response-curve oracle} instead assigns each problem one nominal budget level, including a zero option, subject to the same total budget, and maximizes the expected number of correct answers implied by the response curves. It is therefore a multiple-choice knapsack over budget levels: the oracle chooses not only which problems to fund but how richly to fund each. Because it has offline access to every problem's complete empirical response curve, it serves as an empirical best-allocation reference.

Let $\mathrm{Contest}$ and $\mathrm{Oracle}$ denote their respective
average scores. We report
\[
\Delta_{\mathrm{RR}}=\mathrm{Oracle}-\mathrm{Contest},
\qquad
\mathrm{GapRatio}
=\frac{\Delta_{\mathrm{RR}}}{\mathrm{Oracle}},
\]
with the latter defined when $\mathrm{Oracle}>0$.
A smaller gap means that the contest run realizes more of the
single-problem competence recorded in the response curves.
Appendix~\ref{app:oracle_knapsack} gives the full replay construction and its limitations.

\paragraph{Allocation diagnostics.}
At the outcome level, we define an \emph{oracle-selected miss} as a problem selected by the response-curve oracle but missed in the actual contest. For each oracle-selected miss with sufficient evidence, we assign exactly one primary cause: \emph{never attempted}, \emph{attempted too late}, \emph{stopped after partial progress}, \emph{spent budget elsewhere}, \emph{misread tool feedback}, \emph{wrong format or incomplete finalization}, or \emph{genuinely unsolved}. Misses with insufficient evidence are excluded from the reported cause shares.

Separately, we annotate online adaptation at the trajectory level. We record whether the model observes task-relevant evidence, makes a substantive strategy update in response to a new observation, whether that update is resource-rational with respect to progress, cost, expected value, or remaining budget, and whether the enacted update nevertheless fails to resolve the targeted shortfall. Initial planning and bookkeeping-only changes do not count as strategy updates.

Appendices~\S\ref{app:budget_pressure} and~\S\ref{app:qualitative_cases} provide additional diagnostics and qualitative cases, while Appendix~\S\ref{app:trajectory-annotation} gives the full annotation rubric, denominators, evidence requirements, and decision-regret definitions.

\section{Benchmarking SOTA LLMs}
\label{sec:main_results}
We benchmark eight recent frontier LLMs: DeepSeek-V4-Chat,
DeepSeek-V4-Reasoner, DeepSeek-V4-Pro~\citep{deepseekai2026deepseekv4},
Qwen3.7-Max~\citep{qwen37}, GLM-5.2~\citep{zai2026glm52},
Hy-3~\citep{tencent2026hy3}, GPT-5.5~\citep{openai2026gpt55}, and Claude-Opus-4.8~\citep{anthropic2026claudeopus48}. Evaluation spans both settings, three domains, and two budget pressures. Table~\ref{tab:main-results-domain-columns} reports six flagship models; complete eight-model results appear in
Appendix~\ref{app:full_result_breakdown}. Model interfaces and inference-time configurations are in Appendix~\ref{app:model_card_config}.
\providecommand{\best}{\cellcolor[HTML]{B8E7E1}}
\providecommand{\second}{\cellcolor[HTML]{D4FAFC}}

\definecolor{TableStripe}{HTML}{F1F4F7}
\newcommand{\tablemodellogo}[2]{%
  \raisebox{-0.16em}{%
    \includegraphics[height=1.05em,keepaspectratio]{figs/logos/#1}%
  }%
  \hspace{0.28em}#2%
}

\begin{table*}[!t]
\centering
\scriptsize
\setlength{\tabcolsep}{3.8pt}
\renewcommand{\arraystretch}{1.12}

\caption{
\textbf{Main results on \bench{} with domains as columns.}
For each domain, Contest and Oracle report the average number of correct
answers per six-problem suite. Gap Ratio denotes
$\Delta_{\mathrm{RR}}/\mathrm{Oracle}$, where
$\Delta_{\mathrm{RR}}=\mathrm{Oracle}-\mathrm{Contest}$.
Lower Gap Ratio is better. AR denotes abstract reasoning.
The lowest Gap Ratios are shown in \colorbox[HTML]{B8E7E1}{\textbf{bold}}, and the second-lowest in \colorbox[HTML]{D4FAFC}{\underline{underline}}.
}
\label{tab:main-results-domain-columns}

\resizebox{\textwidth}{!}{%
\begin{tabular}{@{}ll *{9}{c}@{}}
\toprule

\multirow{2}{*}{\textbf{Model}}
& \multirow{2}{*}{\textbf{Regime}}
& \multicolumn{3}{c}{\textbf{Math}}
& \multicolumn{3}{c}{\textbf{Code}}
& \multicolumn{3}{c}{\textbf{AR}} \\

\cmidrule(lr){3-5}
\cmidrule(lr){6-8}
\cmidrule(l){9-11}

&
& \textbf{Contest}
& \textbf{Oracle}
& \textbf{Gap Ratio}
& \textbf{Contest}
& \textbf{Oracle}
& \textbf{Gap Ratio}
& \textbf{Contest}
& \textbf{Oracle}
& \textbf{Gap Ratio} \\

\midrule


\multicolumn{11}{l}{\textbf{\textit{Tool-Free}}} \\

&
$\rho{=}0.2$
& 2.34 & 3.94 & 40.61\%
& 1.38 & 3.12 & 55.77\%
& 2.24 & 4.02 & 44.28\% \\

\multirow{-2}{*}{%
  \tablemodellogo{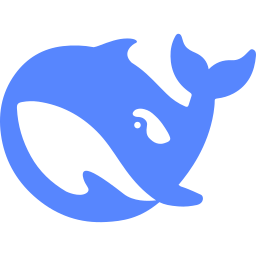}{DeepSeek-V4-Pro}%
}
& $\rho{=}0.8$
& 3.94 & 5.08 & 22.44\%
& 3.06 & 4.62 & 33.77\%
& 2.90 & 4.64 & 37.50\% \\

\rowcolor{TableStripe}
&
$\rho{=}0.2$
& 3.54 & 3.70 & \best\textbf{4.32\%}
& 1.42 & 2.94 & 51.70\%
& 2.76 & 4.06 & 32.02\% \\

\rowcolor{TableStripe}
\multirow{-2}{*}{%
  \tablemodellogo{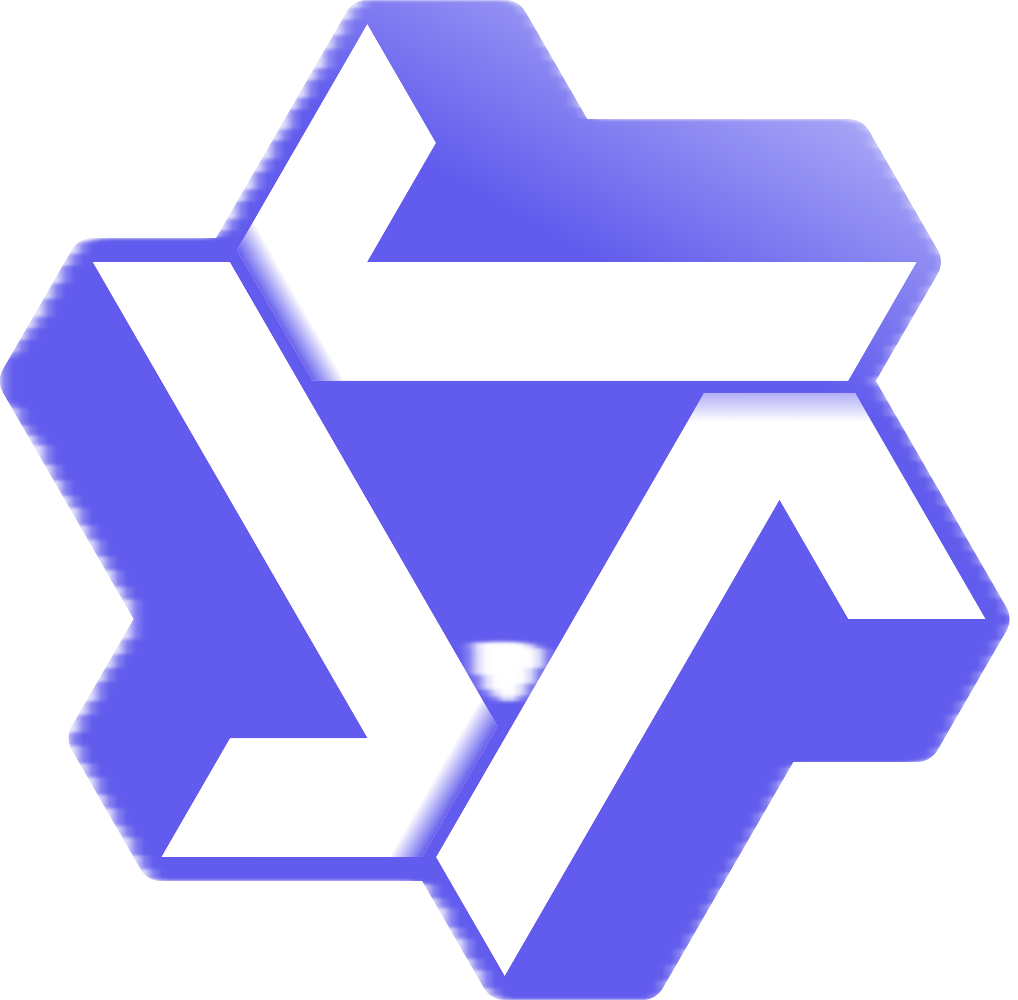}{Qwen3.7-Max}%
}
& $\rho{=}0.8$
& 4.32 & 5.06 & 14.62\%
& 3.44 & 5.24 & 34.35\%
& 3.82 & 4.70 & \best\textbf{18.72\%} \\

&
$\rho{=}0.2$
& 2.32 & 3.08 & 24.68\%
& 0.68 & 1.88 & 63.83\%
& 1.14 & 3.51 & 67.52\% \\

\multirow{-2}{*}{%
  \tablemodellogo{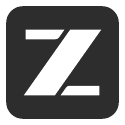}{GLM-5.2}%
}
& $\rho{=}0.8$
& 2.72 & 3.72 & 26.88\%
& 1.72 & 2.08 & \second\underline{17.31\%}
& 1.70 & 4.19 & 59.43\% \\

\rowcolor{TableStripe}
&
$\rho{=}0.2$
& 2.40 & 3.36 & 28.57\%
& 1.20 & 2.28 & 47.37\%
& 1.56 & 3.11 & 49.84\% \\

\rowcolor{TableStripe}
\multirow{-2}{*}{%
  \tablemodellogo{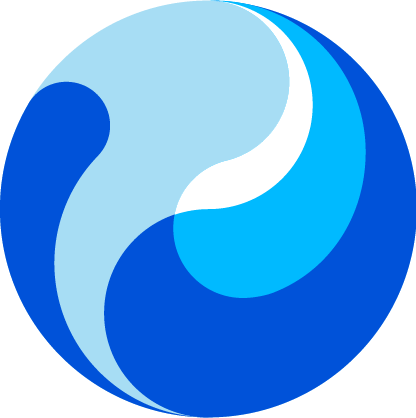}{Hy-3}%
}
& $\rho{=}0.8$
& 3.72 & 4.68 & 20.51\%
& 2.96 & 3.88 & 23.71\%
& 3.16 & 4.16 & \second\underline{24.04\%} \\

&
$\rho{=}0.2$
& 0.72 & 1.28 & 43.75\%
& 0.68 & 1.28 & 46.88\%
& 0.34 & 1.94 & 82.47\% \\

\multirow{-2}{*}{%
  \tablemodellogo{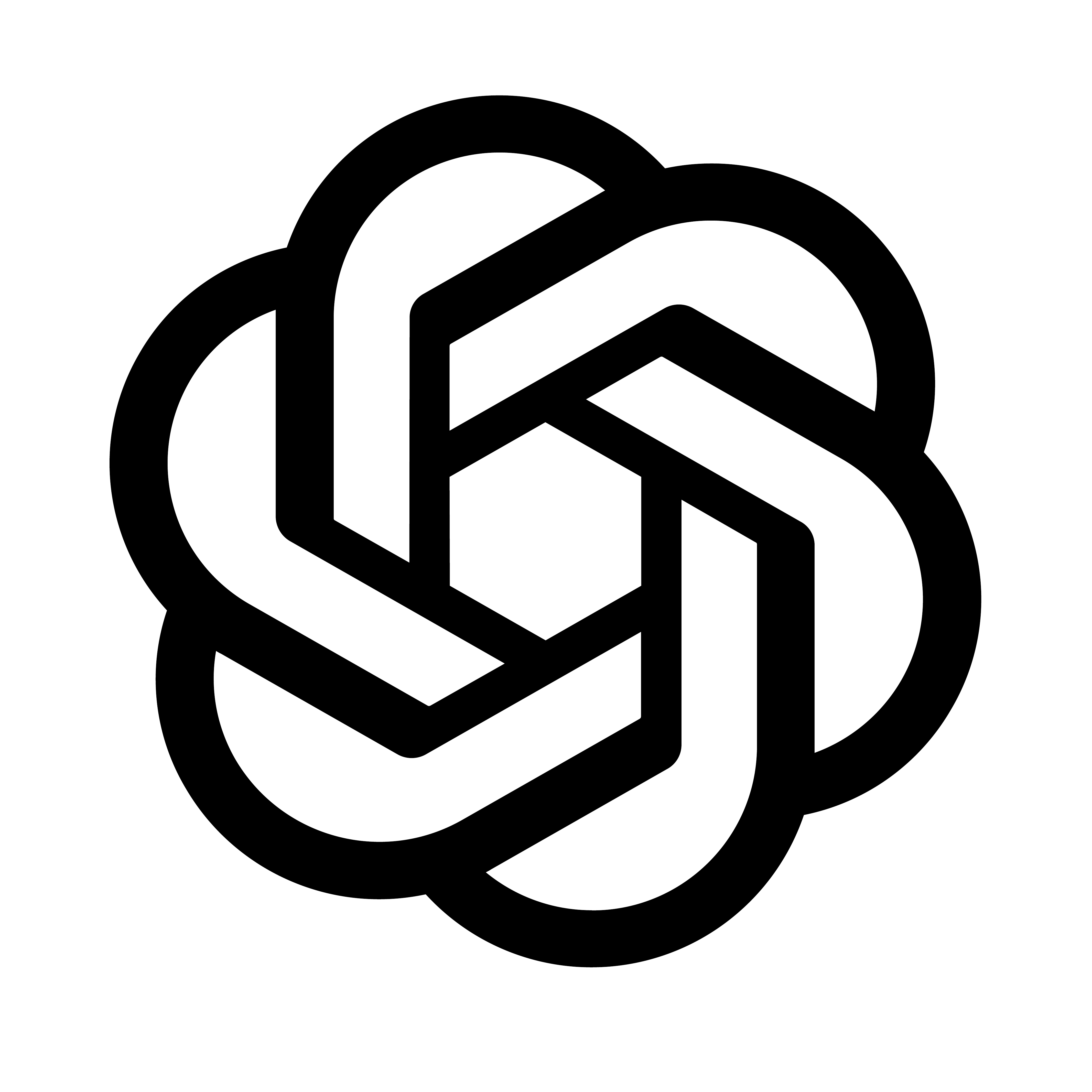}{GPT-5.5}%
}
& $\rho{=}0.8$
& 2.52 & 2.80 & \second\underline{10.00\%}
& 2.60 & 3.12 & \best\textbf{16.67\%}
& 1.58 & 2.75 & 42.55\% \\

\rowcolor{TableStripe}
&
$\rho{=}0.2$
& 3.78 & 4.30 & 12.09\%
& 0.60 & 1.88 & 68.09\%
& 1.84 & 3.11 & 40.82\% \\

\rowcolor{TableStripe}
\multirow{-2}{*}{%
  \tablemodellogo{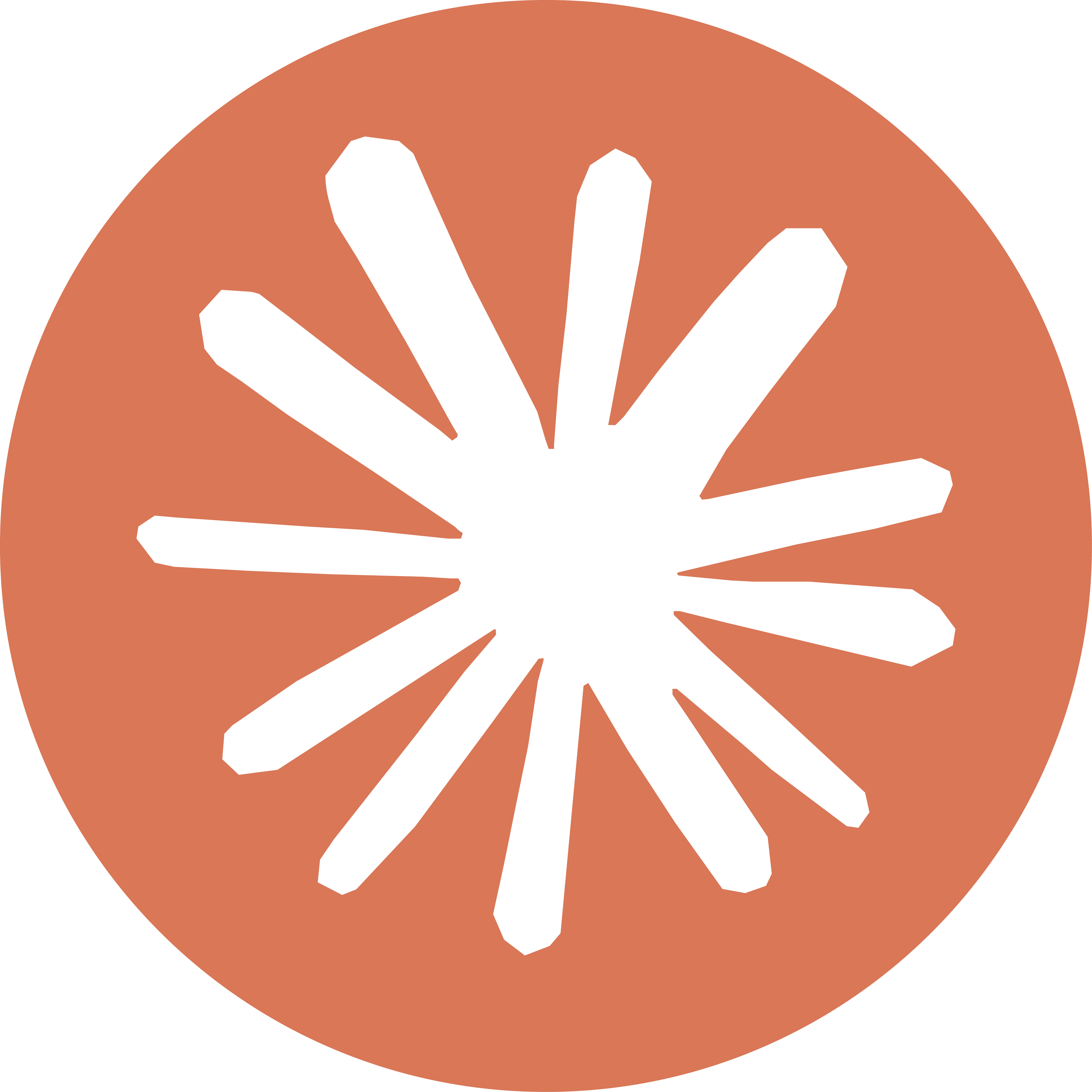}{Claude-Opus-4.8}%
}
& $\rho{=}0.8$
& 4.18 & 4.96 & 15.73\%
& 2.68 & 3.48 & 22.99\%
& 2.52 & 4.15 & 39.21\% \\

\midrule


\multicolumn{11}{l}{\textbf{\textit{Agentic}}} \\

&
$\rho{=}0.2$
& 4.48 & 5.28 & 15.15\%
& 2.60 & 4.90 & 46.94\%
& 4.36 & 5.06 & 13.83\% \\

\multirow{-2}{*}{%
  \tablemodellogo{deepseek.png}{DeepSeek-V4-Pro}%
}
& $\rho{=}0.8$
& 4.48 & 5.40 & 17.04\%
& 4.10 & 5.53 & 25.86\%
& 4.12 & 5.14 & 19.84\% \\

\rowcolor{TableStripe}
&
$\rho{=}0.2$
& 4.64 & 5.14 & 9.73\%
& 2.78 & 2.96 & \second\underline{6.08\%}
& 4.62 & 4.96 & 6.85\% \\

\rowcolor{TableStripe}
\multirow{-2}{*}{%
  \tablemodellogo{qwen.png}{Qwen3.7-Max}%
}
& $\rho{=}0.8$
& 4.96 & 5.34 & 7.12\%
& 4.72 & 5.46 & 13.55\%
& 4.80 & 5.14 & 6.61\% \\

&
$\rho{=}0.2$
& 4.48 & 4.58 & \second\underline{2.18\%}
& 2.02 & 3.90 & 48.21\%
& 3.78 & 4.52 & 16.37\% \\

\multirow{-2}{*}{%
  \tablemodellogo{zhipu.png}{GLM-5.2}%
}
& $\rho{=}0.8$
& 4.16 & 5.12 & 18.75\%
& 4.06 & 5.06 & 19.76\%
& 4.06 & 5.04 & 19.44\% \\

\rowcolor{TableStripe}
&
$\rho{=}0.2$
& 1.80 & 3.38 & 46.75\%
& 1.00 & 3.00 & 66.67\%
& 1.22 & 3.14 & 61.15\% \\

\rowcolor{TableStripe}
\multirow{-2}{*}{%
  \tablemodellogo{hunyuan.png}{Hy-3}%
}
& $\rho{=}0.8$
& 1.56 & 4.16 & 62.50\%
& 1.38 & 3.56 & 61.24\%
& 0.80 & 4.32 & 81.48\% \\

&
$\rho{=}0.2$
& 3.64 & 4.96 & 26.61\%
& 1.92 & 2.00 & \best\textbf{4.00\%}
& 1.80 & 4.92 & 63.41\% \\

\multirow{-2}{*}{%
  \tablemodellogo{gpt.png}{GPT-5.5}%
}
& $\rho{=}0.8$
& 4.40 & 5.32 & 17.29\%
& 4.80 & 5.53 & 13.25\%
& 4.72 & 5.24 & 9.92\% \\

\rowcolor{TableStripe}
&
$\rho{=}0.2$
& 4.46 & 4.46 & \best\textbf{0.00\%}
& 2.24 & 3.56 & 37.08\%
& 3.77 & 3.88 & \best\textbf{2.84\%} \\

\rowcolor{TableStripe}
\multirow{-2}{*}{%
  \tablemodellogo{claude_model.png}{Claude-Opus-4.8}%
}
& $\rho{=}0.8$
& 4.85 & 5.02 & 3.39\%
& 4.52 & 5.08 & 11.02\%
& 5.12 & 5.34 & \second\underline{4.12\%} \\

\bottomrule
\end{tabular}%
}
\end{table*}

\subsection{Overall Results}
Across the 72 reported model--setting--pressure--domain cells in Table~\ref{tab:main-results-domain-columns}, the offline response-curve oracle matches or exceeds the contest score in every cell and is strictly higher in 71, revealing empirical headroom between observed per-task successes and realized shared-budget performance. The agentic setting has a lower Gap Ratio than the tool-free setting in 27 of 36 matched comparisons, while increasing \(\rho\) from \(0.2\) to \(0.8\) lowers it in 23 of 36. These results indicate that resource-rational allocation remains challenging even with interactive feedback and substantially looser relative budgets.

Although recent frontier LLMs have made substantial progress in long-context processing, we still investigate whether the observed gap could be attributable to the increased context length. We run a target-in-suite diagnostic on two models across both settings and all three domains. All six problems are shown, but only one is designated for solution, using the largest single-problem budget in our response-curve evaluation. Across the resulting 12 model--setting--domain cells, nine show point-estimated losses of at most 5 percentage points relative to problem-matched single-problem controls, with an unweighted descriptive mean control--stress difference of 1.1 percentage points (Appendix~\ref{app:context-stress}).

\subsection{Tool-free Reasoning}
\paragraph{Tool-Free Contest Scores Fall below the Oracle.}
\label{subsec:acc_vs_replay}

\begin{figure*}[t]
    \centering
    \includegraphics[width=\textwidth]{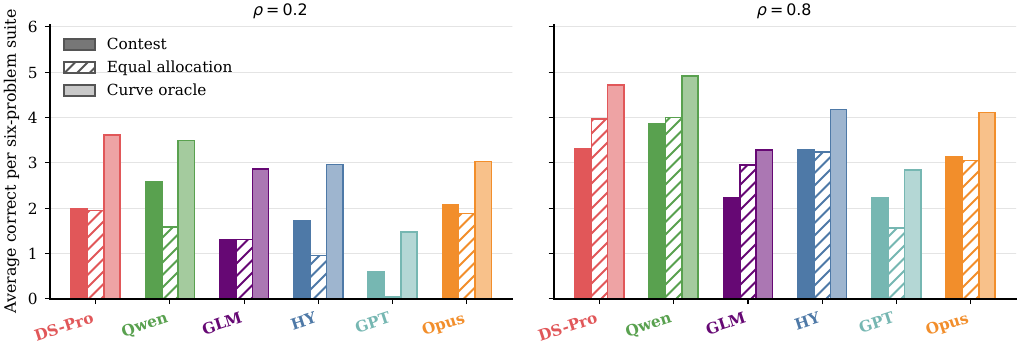}
    \caption{
    Tool-free contest performance, equal-allocation replay, and response-curve
    oracle under strong (\(\rho=0.2\)) and moderate (\(\rho=0.8\)) budget
    pressure. Results are averaged equally over mathematics, competitive
    programming, and abstract reasoning. The unit is average correct answers
    per six-problem suite.
    }
    \label{fig:acc_vs_replay}
\end{figure*}

A model can solve a problem in a per-task run and still miss it when six problems share one budget. Figure~\ref{fig:acc_vs_replay} compares the contest score with the equal-allocation replay and the response-curve oracle defined in Section~\ref{subsec:metrics}.  

The oracle exceeds the contest score for all six models at both pressure levels, with a mean gap of \(1.16\) correct answers across the 12 model--pressure pairs. Because the oracle reuses observed per-task successes under the same total budget, this gap shows that a different allocation of the same budget supports more correct answers; it quantifies offline empirical headroom rather than the performance of an executable policy.  

Equal allocation provides a second test. At \(\rho=0.8 \), it
outperforms the contest for four of the six models: DeepSeek-V4-Pro, Qwen3.7-Max, GLM-5.2, and Claude-Opus-4.8. This fixed
uniform split therefore exposes cases where the model's own
allocation reduces performance. Under strong pressure, however,
the equal share can become extremely small. For GPT-5.5 at
$\rho=0.2$, the per-problem replay thresholds are only 38, 47,
and 6 output tokens for Math, Code, and AR, respectively. Since
a problem counts only when an observed correct single-problem
attempt finishes within the corresponding realized-cost threshold,
the equal-allocation score is near zero. This case also illustrates
the limitation of a fixed equal split under severe budget pressure.

\paragraph{Later Positions Receive Fewer Valid and Correct Answers.}
\label{subsec:position_effect}

\begin{figure}[!b]
    \centering
    \includegraphics[width=\columnwidth]{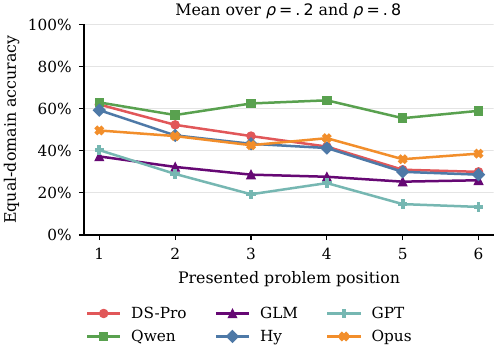}
    \caption{Tool-free accuracy by presented position, averaged over strong and
    moderate budget pressure. Means are unweighted over available domains, with
    missing domains omitted. Accuracy declines from position 1 to position 6 in
    every model--pressure series.}
    \label{fig:position_effect}
\end{figure}

Figure~\ref{fig:position_effect} shows that position 6 has lower accuracy than position 1 in all 12 underlying model--pressure series. Appendix Figure~\ref{fig:judge_position} provides a corresponding trajectory diagnostic: the answered rate falls and the judged budget-truncation rate rises from the first to the last position in all 12 series.  

Together, the oracle, equal-allocation, and position diagnostics indicate that tool-free models do not fully translate observed per-task successes into suite-level reward. Presentation order is randomized independently of difficulty, so the position gradient points to one mechanism: spending follows the order in which problems arrive rather than their worth. Under strong pressure, the first position absorbs 20.5--52.9\% of attributed output, compared with 16.9--23.5\% under moderate pressure. Five of the six models show stronger early-position concentration under \(\rho=0.2\), with Qwen as the exception.  

\subsection{Agentic Reasoning}
\label{subsec:agentic-reasoning}

The agentic setting lets a model probe problems, observe tool feedback, track
the remaining budget, and reallocate resources mid-contest, creating an
opportunity for online adaptation. We ask whether models make substantive
strategy updates after observing difficulty or feedback, and what causes the
remaining oracle gaps.
\begin{figure}[!t]
    \centering
    \includegraphics[width=\columnwidth,height=0.78\textheight,keepaspectratio]{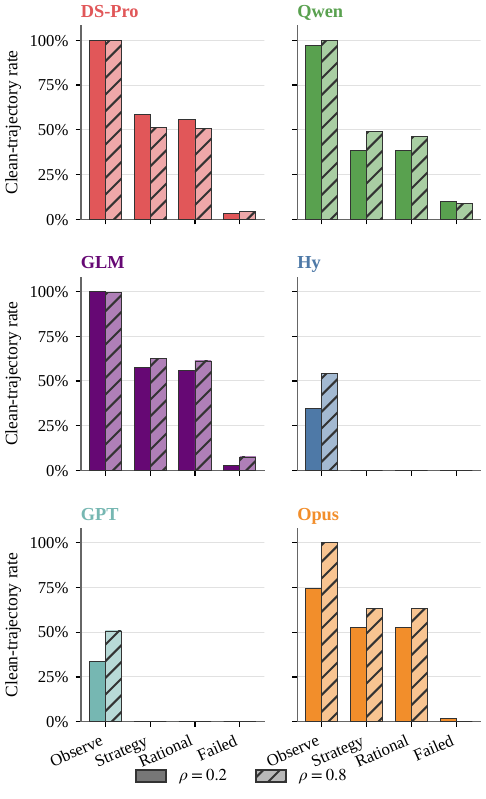}
    \caption{Online adaptation in the agentic setting.
    Rates are equal-domain macros over clean trajectories;
    the annotation rubrics are provided in Appendix~\ref{app:trajectory-annotation}.}
    \label{fig:agentic_funnel}
\end{figure}

\begin{figure*}[!t]
    \centering
    \includegraphics[width=\textwidth,height=0.82\textheight,keepaspectratio]{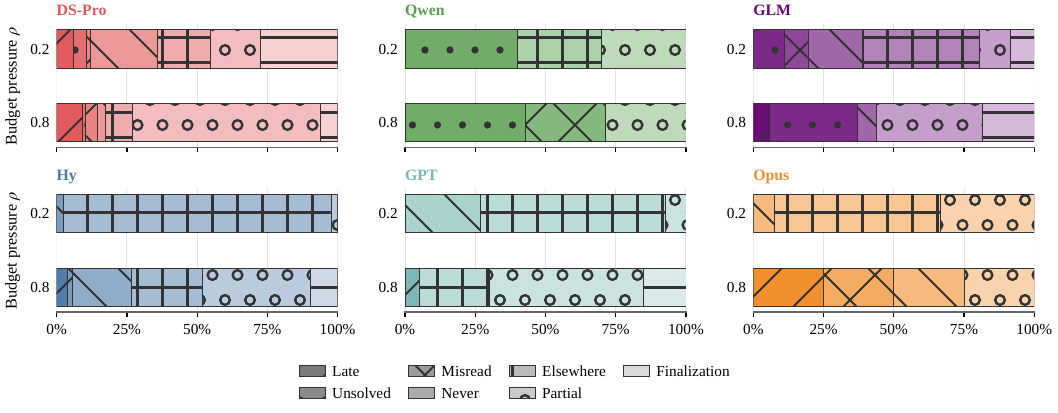}
    \caption{
    Primary causes of positive oracle gaps across models and budget pressures.
    Segment widths show equal-domain macro shares. Budget spent elsewhere
    dominates for several models under strong pressure, and partial progress
    becomes the largest category for most models under moderate pressure.
    }
    \label{fig:agentic_oracle_miss}
\end{figure*}

\paragraph{Online Strategy Changes Remain Limited.}
Figure~\ref{fig:agentic_funnel} shows a clear drop from observing the
environment to changing the allocation strategy. DS-Pro, Qwen, GLM, and Opus
make substantive strategy updates in only \(38.5\%\) to \(63.4\%\) of their
equal-domain macro trajectories, and Hy and GPT make none. Many trajectories observe tool feedback yet
continue with the existing strategy, falling short of the intended online budget reallocation across problems.

\paragraph{Failure Modes Shift with Budget Pressure.}
Figure~\ref{fig:agentic_oracle_miss} summarizes the primary causes of
oracle-selected misses across models and budget pressures. Each oracle-selected miss receives one primary cause
(Appendix~\ref{app:trajectory-annotation}), and which cause dominates
shifts with pressure.

Under strong pressure ($\rho=0.2$), the largest category for most
models is \emph{spent budget elsewhere}: the shared budget went to
other problems, so allocation across problems is what fails.
This category accounts for 42\%, 95\%, 66\%, and 59\% for GLM,
Hy, GPT, and Opus, respectively. Qwen is instead dominated by
\emph{attempted too late}(40\%), while DS-Pro shows a more dispersed
pattern, led by \emph{finalization} (27\%) and \emph{misread tool feedback} (24\%).

Under moderate pressure ($\rho=0.8$), the largest category becomes
\emph{stopped after partial progress} for DS-Pro, GLM, Hy, and GPT,
accounting for 67\%, 38\%, 38\%, and 55\%, respectively. In these
cases, the model opens a problem it has already solved in isolation
and abandons it before submitting an answer. Qwen remains dominated
by \emph{attempted too late} (43\%), while Opus shows an even four-way
split (25\% each).

Because the oracle selects only problems whose observed isolated
costs jointly fit the shared budget, such stops indicate
under-investment rather than a capability limit. Misallocation
therefore persists at $\rho=0.8$ and shifts from \emph{which}
problems receive budget to \emph{how much} they receive.

\subsection{The Allocation Gap Is Not Explained by General Capability}
\label{subsec:effectiveness}

The Epoch Capabilities Index (ECI)~\citep{ho2025rosetta} stitches over fifty
benchmarks into a single capability scale, making it the closest available
summary of what current evaluations measure. We compare it with Gap Ratio,
which is itself a within-model quantity because the oracle is built from each
model's own isolated successes; we ask whether higher general capability still predicts better allocation.

Capability distance carries no information about allocation quality. The four
model pairs within \(3\) ECI points differ by \(9.6\)--\(21.0\) percentage
points in mean absolute Gap Ratio across the twelve cells, and the six pairs
more than \(4\) points apart differ by \(12.0\)--\(19.2\) points: the same spread.
Orderings invert as well; in tool-free Math at \(\rho{=}0.2\) the highest-ECI
model has the largest Gap Ratio (\(43.75\%\)) and a mid-ranked model the
smallest (\(4.32\%\); see Table~\ref{tab:eci_discriminant} in Appendix~\ref{app:full_result_breakdown}).
Over the capability range we can test, then, a composite of over fifty
benchmarks does not predict how well a model spends a shared budget: \bench{}
exposes a dimension that existing evaluations leave uncovered.

\section{Recovering the Allocation Gap}
\label{sec:recovering-allocation-gap}

Section~\ref{sec:main_results} establishes that the allocation gap is broad across the evaluated models. We now ask a narrower question: can a lightweight online scheduling constraint recover part of that gap? We use three models as a diagnostic subset spanning distinct regimes.  DeepSeek-V4-Pro is a strong solver across mathematics, code, and abstract reasoning whose Code result still leaves a substantial gap to the response-curve oracle, making it a useful reference for recoverability. GLM-5.2 is domain sensitive: its agentic gaps and its response to intervention vary sharply by domain, so it tests whether an online scheduler must adapt to the kind of evidence a task provides. Hy-3 is a high-gap boundary case, with the largest mean agentic Gap Ratio across the three domains and no substantive strategy updates under the conservative trajectory rubric of Section~\ref{sec:main_results}, so it tests whether external scheduling can compensate for weak internal adaptation.  We compare the contest reference with two online interventions while holding the agentic loop fixed. Neither intervention reveals answer labels, difficulty labels, response curves, or oracle-selected problems; both use only runtime-visible coverage, budget, and candidate-state information. Strategy A enforces initial coverage: before spending a second paid step on one problem, the agent must give every visible problem a focused probe. Strategy B adds a lightweight verification gate that, once coverage is complete, limits repeated paid checks on a stable candidate unless the candidate is revised. All runs use the agentic setting under strong budget pressure ($\rho=0.2$); implementation details are given in Appendix~\ref{app:online_scheduler_directives}.  

\paragraph{Online scheduling is useful but domain dependent.}
An intervention beats the contest reference in six of the nine model--domain
cells, but no policy dominates: the contest reference, A, and B each take the
best non-oracle score in three rows (Table~\ref{tab:online_scheduler_directives}).
The only regularity is by domain rather than by model. Code improves under both
interventions for every model, whereas Math and AR are model dependent and
scheduling sometimes hurts. We attribute this to the feedback each domain
provides at runtime: compilation and tests make Code probes informative, so
coverage exposes which problems deserve more budget, while the weaker signals
in Math and AR let shallow checks consume budget without indicating where more
reasoning will pay off. Scheduling helps most when runtime feedback tracks the value of more computation.

\paragraph{More complex directives do not guarantee better allocation.}
Adding verification changes the effect of coverage rather than uniformly improving it: B trails A in every Code row, leads in every AR row, and splits in Math. The gate helps when checks resolve uncertainty and interferes when they only consume budget. The directional split argues against treating directive complexity as a monotone design axis.

Hy-3 marks the limit of external scheduling: its best non-oracle
scores stay far below the oracle in all domains, and its Code gain
does not extend to Math or AR. Static directives redirect
computation but do not replace the model's own ability to
interpret progress and revise its plan.

\begin{table}[!t]
    \centering
    \small
    \setlength{\tabcolsep}{4.5pt}
    \renewcommand{\arraystretch}{1.12}

    \caption{Online scheduler results for DeepSeek-V4-Pro, GLM-5.2, and Hy-3
    in the Agentic setting at $\rho=0.2$. Bold marks the best non-oracle score
    in each model--domain row.}

    \label{tab:online_scheduler_directives}

    \resizebox{\columnwidth}{!}{%
    \begin{tabular}{@{}llcccc@{}}
        \toprule

        \textbf{Model}
        & \textbf{Domain}
        & \multicolumn{4}{c}{\textbf{Score}} \\
        \cmidrule(lr){3-6}

        &
        &
        \makecell[c]{\textbf{Contest}\\\textbf{Reference}}
        & \textbf{A}
        & \textbf{B}
        & \textbf{Oracle} \\

        \midrule

        \multirow{3}{*}{DeepSeek-V4-Pro}
        & Math
        & 4.48
        & 4.56
        & \textbf{4.90}
        & 5.28 \\

        & Code
        & 2.60
        & \textbf{4.10}
        & 3.86
        & 4.90 \\

        & AR
        & 4.36
        & 4.32
        & \textbf{4.72}
        & 5.06 \\

        \midrule

        \multirow{3}{*}{GLM-5.2}
        & Math
        & \textbf{4.48}
        & 4.04
        & 4.20
        & 4.58 \\

        & Code
        & 2.02
        & \textbf{2.92}
        & 2.66
        & 3.90 \\

        & AR
        & 3.78
        & 4.10
        & \textbf{4.44}
        & 4.52 \\

        \midrule

        \multirow{3}{*}{Hy-3}
        & Math
        & \textbf{1.80}
        & 1.64
        & 1.58
        & 3.38 \\

        & Code
        & 1.00
        & \textbf{1.56}
        & 1.36
        & 3.00 \\

        & AR
        & \textbf{1.22}
        & 0.48
        & 0.70
        & 3.14 \\

        \bottomrule
    \end{tabular}%
    }
\end{table}

These results motivate an inner scheduling policy implemented through training. The policy can be trained to condition its decisions on the task domain, the quality of current progress, and the remaining budget, and to select when to cover, continue, switch, verify, or stop. Such a policy could preserve
domain-specific behavior and reduce interference with correct model decisions.
Training and evaluating this policy remain future work. The intervention study
highlights the limited cross-domain generality of a single fixed directive.

\section{Conclusion}

We introduced \bench{}, a benchmark for resource-rational reasoning under a shared computational budget. Existing benchmarks evaluate problems independently. \bench{} instead places several problems under one shared budget. It compares contest performance with what the same model has already demonstrated on single problems. The benchmark covers mathematics, competitive programming, and abstract reasoning. In both tool-free and agentic settings, current models leave a persistent gap to their response-curve oracle. Single-problem competence therefore does not guarantee effective allocation across problems.

The gap reflects failures in both where and how much computation
is allocated. Under strong pressure, models often spend the shared
budget on other problems, while under moderate pressure they more
often stop after partial progress. Online adaptation and lightweight external scheduling can alleviate this gap, but they do not close it. Broader coverage helps when runtime feedback is informative, but can hurt when shallow probes provide weak evidence. Resource-rational reasoning therefore requires more than additional computation. It requires
policies that decide when to continue, when to switch,
and where to spend the remaining budget. Learning such an
internal allocation policy is a promising future direction.

\FloatBarrier

\bibliographystyle{aaai2026}
\bibliography{reference}

\clearpage
\appendix
\setcounter{topnumber}{4}
\setcounter{dbltopnumber}{4}
\renewcommand{\topfraction}{0.95}
\renewcommand{\dbltopfraction}{0.95}
\renewcommand{\textfraction}{0.05}
\renewcommand{\floatpagefraction}{0.75}
\renewcommand{\dblfloatpagefraction}{0.75}
\raggedbottom
\makeatletter
\setlength{\@fptop}{0pt}
\setlength{\@fpbot}{0pt plus 1fil}
\setlength{\@dblfptop}{0pt}
\setlength{\@dblfpbot}{0pt plus 1fil}
\makeatother
\enableappendixcontents
\section*{Appendix Contents}
\appendixcontents
\section{Data Sources}
\label{app: data_source}
\textbf{Omni-MATH}~\citep{gao2025omni} is an Olympiad-level mathematics benchmark designed to evaluate advanced mathematical reasoning in large language models. The dataset contains 4,428 text-only competition problems, collected from contest pages and AoPS-style sources and further verified through human annotation. It spans more than 33 mathematical subdomains and over 10 difficulty levels, enabling fine-grained analysis across both topic areas and problem complexity. In addition to standard answer checking, Omni-MATH provides model-based evaluation through GPT-4o and the open-source Omni-Judge verifier for assessing open-ended mathematical solutions.  

\textbf{MathNet}~\citep{alshammari2026mathnet} is a large-scale multilingual and multimodal benchmark for Olympiad-level mathematical reasoning and retrieval. Its main corpus comprises 30,676 expert-authored problems with solutions, covering 47 countries, 16 languages, and diverse mathematical domains. Beyond direct problem solving, MathNet introduces retrieval-oriented tasks based on mathematically equivalent or structurally similar problem pairs, supporting evaluation of math-aware retrieval and retrieval-augmented problem solving. This design makes it suitable for assessing not only generative mathematical reasoning, but also whether models can recognize deep structural similarity across notation, language, and modality.

\textbf{LiveCodeBench Pro}~\citep{zheng2025livecodebench} is a continuously updated competitive-programming benchmark for evaluating code-centric reasoning under reduced data-contamination risk. The benchmark contains 584 high-quality problems, collected before April 25, 2025, from top-tier contests including Codeforces, ICPC, and IOI, while excluding more contamination-prone sources such as LeetCode. Each problem is annotated by Olympiad medalists with algorithmic skill tags and cognitive-focus labels such as knowledge-heavy, observation-heavy, and logic-heavy. Together with Codeforces-style difficulty tiers and Elo-based evaluation, LiveCodeBench Pro provides both aggregate performance measures and fine-grained diagnostics of model failures in competitive programming.  

\textbf{Reasoning Gym}~\citep{stojanovski2025reasoning} is a procedural collection of reasoning environments for reinforcement learning with verifiable rewards, rather than a fixed static dataset. It provides over 100 data generators and automatic verifiers across domains such as algebra, arithmetic, computation, cognition, geometry, graph theory, logic, and games. Each environment can generate virtually unlimited problem instances with controllable difficulty and algorithmic scoring, making it useful for both systematic evaluation and curriculum-style reinforcement learning. This design allows researchers to study reasoning ability under adjustable complexity while avoiding the limitations of finite, memorization-prone benchmark sets.

\section{Protocol for Thinking Models}
\label{app:thinking_models}

For the reported tool-free two-stage runs of thinking models, we separate
budgeted problem solving from answer finalization. This protocol is not used in
the agentic setting, where models interact with tools and write final artifacts
directly.

\paragraph{Stage 1: reasoning and drafting.}
Stage~1 is the problem-solving stage. The model receives either one problem or
the full six-problem suite and runs with thinking or reasoning enabled. A
single-problem run uses its budget for that problem; a contest run allocates one
shared budget across the six problems. In both settings, the model is instructed
to preserve candidate final answers, partial results, or complete candidate
programs whenever possible.

In all three domains, the run-specific budget \(B^\rho_{m,d}\), where
\(\rho\in\{0.2,0.8\}\), is applied to Stage~1. Stage~2 does not draw on this
budget. Configured caps and provider-reported token usage are recorded
separately for both stages.

\paragraph{Stage 2: finalization.}
Stage~2 runs with thinking disabled and receives no reference answers, hidden
tests, verifier outcomes, or correctness feedback. Its role is to convert the
stopped Stage~1 output into the required answer format rather than to serve as
an independently evaluated solve attempt.

For the reported mathematics and abstract-reasoning contests, Stage~2 is a
trace-only finalizer. It is not given the original problem statements. It
receives only the available Stage~1 reasoning content and visible output. It is
instructed to extract answers supported by that material and to return a
missing-answer marker when the available information is insufficient. This is
a prompt-level restriction. We audit compliance manually, as reported below.

For the reported coding thinking-model contests, Stage~2 is a trace-only
finalizer under the same rule. It is not given the original problem statements.
It receives the available Stage~1 reasoning content and visible output and
prints the complete C++ programs submitted to the verifier. It is instructed to
produce only programs supported by the Stage~1 trace.

\begin{table*}[t]
\centering
\small
\setlength{\tabcolsep}{5pt}
\begin{tabular}{p{0.20\textwidth}p{0.34\textwidth}p{0.36\textwidth}}
\toprule
Domain & Stage~2 input & Finalization constraint \\
\midrule
Mathematics / AR trace-only finalizer
& Stage~1 reasoning and visible output; no original problem statements
& Extracts trace-supported answers and returns a missing-answer marker when the trace is insufficient \\
Coding trace-only finalizer
& Stage~1 reasoning and visible output; no original problem statements
& Prints complete trace-supported C++ programs, or \texttt{MISSING} when the trace is insufficient \\
\bottomrule
\end{tabular}
\caption{Two-stage protocols used by the reported Tool-free thinking-model
contests. In all three domains, Stage~1 receives the run-specific budget
\(B^\rho_{m,d}\) and Stage~2 output is logged separately rather than charged to
that budget.}
\label{tab:thinking_model_profiles}
\end{table*}

\paragraph{Budget accounting and judging.}
Budget accounting is uniform across the three domains: in every case the
metered quantity is Stage~1 completion tokens, and Stage~2 output is logged
separately rather than charged. In contests, Stage~1 usage is charged against
the run-specific budget \(B^\rho_{m,d}\). In single-problem response-curve
runs, the charged cost is the configured grid level rather than the realized
usage of the individual run, as specified in
Appendix~\ref{app:oracle_knapsack}.

Two properties prevent the uncharged Stage~2 from acting as extra
problem-solving capacity. It runs with thinking disabled and without the
original problem statements, so it cannot solve a problem that Stage~1 left
unsolved, and it must emit a missing-answer marker, or \texttt{MISSING} for
coding, whenever the trace does not support an answer. Because these
requirements are prompt-level, we record Stage~2 token usage, finish reasons,
and API or parsing failures separately and report them as protocol outcomes.
Stage~2 calls remain subject to ordinary runtime and provider safety limits.
The resulting outputs are graded by the domain-specific pipelines described in
Appendix~\ref{app:answer_judging}.

\paragraph{Manual audit of Stage 2.}
We verified directly that the prompt-level restriction holds in the reported
runs. The audit covers the five models with a separate reasoning channel, at
both pressure levels and in all three domains, giving
\(5\times 2\times 3=30\) model--pressure--domain cells; from each cell we drew
\(10\) Stage~2 calls at random, for \(300\) audited calls. For each call we
paired the Stage~2 output with its own Stage~1 trace and required every emitted
answer, and for coding every emitted program, to correspond to a candidate
answer or candidate program already present in that trace; we then checked the
Stage~2 output for derivation steps or algorithms that the trace does not
contain. No audited call re-solved a problem. Every Stage~2 output was
attributable to Stage~1 material, and none introduced a new derivation or a new
algorithm. Stage~2 thus behaves as the formatting channel the protocol
specifies, and its uncharged tokens do not supply problem-solving capacity that
would affect comparability across budgets.

\paragraph{Interpretation.}
The two-stage protocol reduces a formatting-related confound in reasoning-mode
evaluation. Under a small output cap, a model may consume most of its allowance
in the reasoning channel before emitting parseable final answers. Separating
reasoning and drafting from finalization avoids requiring every answer to be
printed before Stage~1 stops, so in all three domains the budget measures
reasoning and drafting rather than transcription. Models without a separate
reasoning channel are evaluated in a single stage, so their final-answer tokens
are charged to \(B^\rho_{m,d}\). Finalization failures remain visible: Stage~2
cap hits, API errors, parser failures, and missing outputs are retained as
separately logged protocol outcomes.
\section{Answer Parsing and Judging}
\label{app:answer_judging}

This appendix specifies how model outputs are parsed into problem-level
answers and how those answers are judged. For each domain and setting, we
define the required answer artifact, the parser that extracts per-problem
answers, the judge or verifier that assigns correctness, and the treatment of
missing, malformed, or unparsable outputs. Correctness is assigned at the
problem level and then aggregated to the contest level. Behavioral labels used
in trajectory analyses are diagnostic only and are not used to compute
accuracy, oracle scores, or contest rewards.
Table~\ref{tab:answer_contracts_and_judges}
summarizes the answer contract, final artifact, and correctness judge for each
domain.

\begin{table*}[t]
\centering
\scriptsize
\setlength{\tabcolsep}{4pt}
\renewcommand{\arraystretch}{1.08}
\begin{tabular}{p{0.18\textwidth}p{0.25\textwidth}p{0.24\textwidth}p{0.25\textwidth}}
\toprule
Domain & Pure-NL answer contract & Agentic final artifact & Correctness judge \\
\midrule
Mathematics
&
\texttt{\textbackslash boxed\{...\}} final answer
&
\texttt{/logs/artifacts/answer.txt}
&
Boxed-answer extraction followed by an offline model-based equivalence judge
\\
Competitive programming
&
Complete standalone C++17 solution, usually in a fenced code block
&
\texttt{/app/solution\_A.cpp}--\texttt{/app/solution\_F.cpp}
&
LightCPVerifier over hidden tests; \textsc{Accepted} is correct
\\
Abstract reasoning
&
\texttt{<answer>...</answer>} final answer
&
\texttt{/logs/artifacts/answer.txt}
&
Reasoning Gym answer extraction and dataset-specific scorers
\\
\bottomrule
\end{tabular}
\caption{Domain-specific answer contracts and correctness judges.
\bench{} does not use one global judge for all domains.}
\label{tab:answer_contracts_and_judges}
\end{table*}

\paragraph{Problem-level scoring.}
Each contest \(c\) contains six problems
\(\{p_{c,i}\}_{i=1}^{6}\). Let
\[
    \mathcal{E}_{\mathrm{parse}}
    =
    \{\mathsf{Missing},\mathsf{Malformed},\mathsf{Unparsable}\}.
\]
Given a model output or final artifact \(y\), the domain- and setting-specific
parser produces one record per problem:
\[
\begin{aligned}
    r_{c,i}
    &=
    \operatorname{Parse}_{d,s}(y,p_{c,i}) \\
    &\in
    \mathcal{A}_{d} \cup \mathcal{E}_{\mathrm{parse}} .
\end{aligned}
\]
Here, \(d\) is the domain, \(s\) is the evaluation setting, and
\(\mathcal{A}_{d}\) is the domain-specific answer space.
\(\mathsf{Missing}\) means that no eligible answer or final artifact is
present; \(\mathsf{Malformed}\) means that output exists but violates the
required structural contract; and \(\mathsf{Unparsable}\) means that the
output is present but cannot be converted by the parser into a domain-specific
answer record.

If the parser returns an answer \(\hat a_{c,i}\in\mathcal{A}_{d}\), the
domain-specific judge assigns a binary correctness label
\[
    z_{c,i}
    =
    \operatorname{Judge}_{d}(p_{c,i},\hat a_{c,i})
    \in \{0,1\}.
\]
Offline judges may use the domain's reference answer, hidden tests, or
dataset-specific verifier, but those signals are unavailable to the solver
during the run. Parser-error states receive zero credit. If the parser returns
an answer but the downstream judge or verifier fails, times out, or returns an
invalid status, the problem is also assigned zero credit and the failure is
retained as a diagnostic status. When a domain scorer returns a numeric score,
the main accuracy tables binarize full credit.

The contest score is the number of correct answers in the six-problem suite,
\[
    Z_c = \sum_{i=1}^{6} z_{c,i},
\]
and the reported contest-level accuracy is \(Z_c/6\), averaged over contests.
Difficulty-specific accuracies are computed by aggregating the same
problem-level labels within the Easy, Medium, and Hard buckets. The
with-difficulty and without-difficulty variants use the same parsing, judging,
and aggregation pipeline; the only difference is whether difficulty labels are
shown in the prompt.

\paragraph{Tool-free parsing.}
In the tool-free setting, the model produces one non-interactive
completion. In single-problem runs, the parser extracts one answer according to
the domain's canonical answer contract. In contest runs, the parser first
decomposes the completion into six problem-level sections and then applies the
same domain-specific extraction rule to each section.

For coding, single-problem parsing extracts a C++ solution from structured or
fenced code first and falls back to a complete \texttt{int main} program when
available. Contest parsing is stricter: the expected output is divided into
\texttt{Problem A} through \texttt{Problem F} sections, each containing one C++
solution. If section labels are absent, the parser may use an ordered fallback
only when exactly six code blocks are present. Missing sections, duplicate
labels, multiple code blocks in one section, unclosed fences, or malformed
stage outputs are recorded as parser failures and are not submitted for credit.

For mathematics, the requested final answer format is a boxed answer. Contest
outputs are split into six numbered problem sections, with a fallback search
for problem-level final-answer lines. If no answer can be found for a problem,
the judged solution is treated as no answer rather than as a valid mathematical
expression. Correctness is then decided by an offline equivalence judge that
compares the model answer to the reference answer and returns a binary
equivalence decision.

For abstract reasoning, the parser extracts
\texttt{<answer>...</answer>} tags, with a fallback for final-answer lines when
available, and passes the result to the corresponding Reasoning Gym scorer.
Some scorers return numeric values; for the main accuracy tables, full credit
is binarized as \(\mathrm{score}=1.0\).

\paragraph{Two-stage extraction and finalization.}
For reasoning-mode models, Appendix~\ref{app:thinking_models} defines the
two-stage protocol. Here we specify how the Stage~2 output enters the
grading pipeline. In the reported coding thinking-model contests, Stage~2 is a trace-only finalizer. It receives Stage~1 reasoning content and
visible output, but not the original problem statements, and emits one complete
C++17 program or \textsc{Missing} for each of Problems A--F. The requirement to
use only trace-supported material is imposed by the prompt. The resulting Stage~2 programs are
parsed and evaluated independently by LightCPVerifier.

In mathematics and abstract reasoning, Stage~2 is a finalizer over the stopped
scratchpad and visible partial answers, and its output is passed to the
ordinary parser and judge. In every domain, Stage~2 tokens are recorded
separately as formatting overhead and Stage~2 is not treated as an additional
independent solve attempt, although coding finalization carries the stricter
requirement of serializing complete executable programs.
\paragraph{Agentic parsing and final artifacts.}
In the agentic setting, correctness is based on eligible final artifacts after
task completion, timeout, or budget exhaustion, rather than on live judge
feedback. For coding, the agent writes one solution file per contest problem:
\texttt{/app/solution\_A.cpp} through \texttt{/app/solution\_F.cpp}. Each file
is judged independently by LightCPVerifier. If a required solution file is
absent, the corresponding problem is marked as a missing solution and receives
zero credit. Non-\textsc{Accepted} verdicts, including wrong answer,
compilation error, runtime error, timeout, and verifier errors, are recorded as
diagnostic statuses but receive no correctness credit.

For mathematics and abstract reasoning, the agent writes final answers to
\texttt{/logs/artifacts/answer.txt}. The mathematics grader extracts boxed
answers from the relevant problem sections and then applies the offline
equivalence-judging protocol. If no boxed answer is found for a problem, the
problem receives zero credit and contributes to the missing-box rate. The
abstract-reasoning grader extracts \texttt{<answer>...</answer>} tags and uses
Reasoning Gym scorers; missing tags receive zero credit and contribute to the
missing-answer-tag rate. Details of which finalization actions are counted or
free under each agentic policy are given in Appendix~\ref{app:white_list}.
Across all domains, missing artifacts, parser failures, malformed outputs, and
downstream judge or verifier failures receive zero credit and are retained only
as diagnostic statuses.

\paragraph{Separation from behavior labels and oracle replay.}
Correctness labels from the domain-specific judges are the only labels used for
problem accuracy, contest score, and response-curve oracle construction. Behavioral labels from the human trajectory annotation in
Appendix~\ref{app:trajectory-annotation} are used for diagnostics only
and never affect whether a problem is counted correct. Likewise, the response-curve oracle in
Appendix~\ref{app:oracle_knapsack} does not introduce a new judge: it only
reallocates budget over single-problem outcomes that have already been judged
by the domain-specific correctness pipeline.

\section{Evaluation Prompt Templates}
\label{app:prompts}

This section summarizes the prompt templates used in \bench{}. We deliberately
layer each prompt so that reusable instructions form a stable prefix and
run-specific content appears later, increasing opportunities for provider-side
prefix-cache reuse across API calls~\citep{gim2024prompt}. The exact prompts
and scripts are released with the benchmark artifact. No solver-facing prompt
exposes privileged evaluation information, including oracle decisions,
reference answers, hidden-test outcomes, response-curve selections, or judge
feedback.

\begin{table*}[t]
\centering
\small
\begin{tabular}{p{0.18\linewidth}p{0.25\linewidth}p{0.31\linewidth}p{0.17\linewidth}}
\toprule
Domain & Tool-Free final answer & Agentic contest artifact & Judge \\
\midrule
Mathematics & \texttt{Final Answer: \textbackslash boxed\{...\}} &
\texttt{/logs/artifacts/answer.txt} & model-based equivalence judge \\
Code & C++17 code block &
\texttt{/app/solution\_A.cpp}--\texttt{/app/solution\_F.cpp} &
executable verifier \\
Abstract reasoning & \texttt{Final Answer: <answer>...</answer>} &
\texttt{/logs/artifacts/answer.txt} & rule-based verifier \\
\bottomrule
\end{tabular}
\caption{Domain-specific answer-format instructions shown in the prompts. All
contest prompts require one independently parseable answer per problem.}
\label{tab:prompt_contracts}
\end{table*}

\subsection{Tool-free Reasoning}

\paragraph{System and common prompt layer.}
The Tool-Free runs use domain-specific system or instruction prefixes rather
than a single global system prompt. The mathematics single-problem template
uses the system role shown below. Abstract-reasoning and coding templates place
the solver role, output contract, and tool restrictions directly in the user
instruction. Agentic runs additionally receive the native-tool protocol note
in the agentic section below.

\paragraph{Single-problem prompts.}
For single-problem runs, the model receives one problem and must output one
final answer in the domain-specific format in
Table~\ref{tab:prompt_contracts}. The mathematics template uses an
educator-style system prompt and asks the model to place the final answer in
\(\backslash\texttt{boxed}\{\}\). The abstract-reasoning template follows the
Reasoning Gym format and requires the answer inside
\texttt{<answer>...</answer>} tags. The coding template requests exactly one
complete standalone C++17 program in a fenced code block and disallows tools.

\begin{promptbox}[Mathematics Single-Problem Prompt]
{tsinghuapurple}{box:math_single_prompt}
\footnotesize\ttfamily\raggedright
System: You are an experienced educator in the field of MATHEMATICS.\\
User: \{problem\}\\
Please put your final answer in \textbackslash boxed\{\}.
\end{promptbox}

\begin{promptbox}[Abstract-Reasoning Single-Problem Prompt]
{tsinghuapurple}{box:ar_single_prompt}
\footnotesize\ttfamily\raggedright
Given a problem, answer by thinking step by step. Once finished, provide only
the final answer in <answer>answer here</answer>.\\
User: \{question\}
\end{promptbox}

\begin{promptbox}[Coding Single-Problem Prompt]
{tsinghuapurple}{box:code_single_prompt}
\footnotesize\ttfamily\raggedright
Solve this competitive programming problem. Return exactly one complete C++17
solution in a fenced cpp code block. Do not use tools. Do not include
explanations before or after the code block.
\end{promptbox}

For budgeted single-problem response curves, the prompt adds a domain-specific
budget sentence such as ``Your total output budget for this problem is limited
to \texttt{\{budget\}} tokens.'' For thinking models, the budget instruction
states that hidden reasoning tokens and final-answer tokens both count toward the response budget. The \texttt{{budget}} placeholder is filled with the five response-curve caps defined in Appendix~\ref{app:oracle_knapsack}.

\paragraph{Contest prompts.}
Mathematics and abstract-reasoning contest prompts share the same structure:
the model receives all six problems, a shared output-token budget, and the
instruction to maximize the number of correct final answers. The prompt asks
the model to scan the full suite, choose problems it expects to solve reliably
with few tokens, and emit answers in a fixed per-problem section format.

\begin{promptbox}[Tool-Free Contest Prompt]
{blue!60!black}{box:pure_nl_contest_prompt}
\footnotesize\ttfamily\raggedright
\setlength{\parindent}{0pt}

\textbf{[SYSTEM PROMPT]}\par
\medskip

You are a problem-solving model. You will be given a contest containing
multiple independent problems.\par
\medskip

Your objective is to maximize the total contest score within a shared
response-token budget. You may choose which problems to attempt and how much
reasoning effort to allocate to each problem.\par
\medskip

Solve the problems using natural-language reasoning only. You cannot use
external tools, execute programs, access the live judge, inspect hidden test
data, or consult reference solutions.\par
\medskip

A contest may be submitted partially. You are not required to solve every
problem. An unsubmitted problem receives zero credit. Do not fabricate an
answer merely to complete all problems.\par
\medskip

Use your internal reasoning efficiently, and place only the requested
submissions in the final response.\par
\bigskip

\textbf{[CONTEST PROMPT]}\par
\medskip

You are given a contest with \{NUM\_PROBLEMS\} independent problems.\par
\medskip

Shared response-token budget: \{TOTAL\_TOKEN\_BUDGET\}\par
\medskip

The token budget is shared across the entire contest rather than allocated
separately to each problem. Allocate your reasoning effort carefully to
maximize the total score.\par
\medskip

Problems:\par
\medskip

===== Problem \{PROBLEM\_ID\_1\}: \{PROBLEM\_TITLE\_1\} =====\par
\{PROBLEM\_STATEMENT\_1\}\par
...\par
\medskip

[REMAINING PROBLEMS]\par
\end{promptbox}

\begin{promptbox}[Tool-Free Contest Prompt (continued)]
{blue!60!black}{box:pure_nl_contest_prompt_requirements}
\footnotesize\ttfamily\raggedright
\setlength{\parindent}{0pt}

Response requirements:\par
\medskip

1. For each attempted problem, provide the final answer using the following
format:\par
\medskip

\{PER\_PROBLEM\_ANSWER\_FORMAT\}\par
\medskip

2. Every submitted answer must satisfy:\par
\medskip

\{ANSWER\_FORMAT\_REQUIREMENTS\}\par
\medskip

3. Clearly associate each submitted answer with its problem identifier.\par
\medskip

4. Problems without a valid answer are treated as unsubmitted and receive
zero credit.\par
\medskip

5. You may submit answers for any subset of the problems. Solving all
\{NUM\_PROBLEMS\} problems is not required.\par
\medskip

6. Do not include a fabricated or placeholder answer for a problem you
cannot solve.\par
\medskip

7. Return all selected submissions in a single final response.

\end{promptbox}

The mathematics answer format is
\(\backslash\texttt{boxed}\{...\}\), and the abstract-reasoning answer format
is \texttt{<answer>...</answer>}. In with-difficulty variants, each problem
block receives an additional difficulty line; without-difficulty variants omit
this line.

\subsection{Reasoning-Model Two-Stage Protocol}

For models with a separate thinking or reasoning channel, we use a two-stage
protocol to separate budgeted reasoning from final formatting. In all three
domains, Stage~1 is the budgeted reasoning stage and Stage~2 is a trace-only
finalizer that does not receive the original problem statements. For
mathematics and abstract reasoning it formats final answers from the stopped
scratchpad; for the reported coding thinking-model contests it must construct
the final C++17 submissions from the Stage~1 reasoning and visible output.

\begin{promptbox}[Thinking-Model Stage 1 Prompt]
{magenta!60!black}{box:thinking_stage1_prompt}
\footnotesize\ttfamily\raggedright
You are stage 1 of a two-stage \{domain\_name\} contest solver. You will
receive \{num\_problems\} problems. Produce a concise scratchpad for a later
answer finalizer. Your reasoning/scratchpad budget is \{reasoner\_budget\}
tokens and includes thinking tokens and visible partial-answer tokens. Work on
as many problems as possible; record candidate final answers or No answer.
\end{promptbox}

\begin{promptbox}[Thinking-Model Stage 2 Finalizer Prompt]
{magenta!60!black}{box:thinking_stage2_prompt}
\footnotesize\ttfamily\raggedright
You are finalizing a stopped reasoning trace. Do not solve from scratch. Use
the scratchpad as the source. If the scratchpad does not contain enough
information for a problem, output No answer. Output only final answers in the
required format.
\end{promptbox}

If a coding Stage~1 trace does not contain enough information to support a
complete standalone program, the finalizer outputs \texttt{MISSING}.

\subsection{Agentic Prompts}

Agentic runs use native terminal tools. The runtime prepends a protocol note
and, for budgeted runs, a budget note to the task instruction. Detailed
counted/free action rules are given in Appendix~\ref{app:white_list}; here we
record the task-level prompt contract.

\paragraph{Common native-tool note.}
The native-tool note instructs the model to act through the provided
\texttt{bash\_command} and \texttt{mark\_task\_complete} tools. It forbids
writing Terminus JSON or XML in assistant message content, requires executable
actions to be sent as native tool calls, and instructs thinking models to keep
reasoning in the hidden reasoning channel when available.

\paragraph{Agentic contest instruction.}
The Math/AR agentic prompt asks the model to solve as many problems as possible
and write final answers to \texttt{/logs/artifacts/answer.txt}. It specifies
the required answer format for each problem and may include a runtime budget
note and tool note before the task text.

\begin{promptbox}[Agentic Contest Prompt]
{green!50!black}{box:agentic_math_ar_prompt}
\footnotesize\ttfamily\raggedright
\setlength{\parindent}{0pt}

\textbf{[SYSTEM PROMPT]}\par
\medskip

You are an autonomous problem-solving agent operating in a tool-enabled
environment. You will be given a contest containing multiple independent
problems.\par
\medskip

Your objective is to maximize the total contest score under a shared
computational budget. You may choose which problems to attempt and how much
budget to allocate to each problem.\par
\medskip

You may use the provided tools to inspect the environment, perform
computations, write solution files, and run local tests. Do not access the
live judge, hidden test data, reference solutions, or other prohibited
evaluation resources.\par
\medskip

A contest may be submitted partially. You are not required to solve every
problem. An unsubmitted problem receives zero credit. Do not fabricate an
answer merely to complete all problems.\par
\medskip

Only actions designated as counted actions consume the shared budget.
Bookkeeping and final-submission actions are governed by the evaluation
runtime.\par
\bigskip

\textbf{[CONTEST PROMPT]}\par
\medskip

You are given a contest with \{NUM\_PROBLEMS\} independent problems.\par
\medskip

Shared counted-action budget: \{TOTAL\_BUDGET\}\par
\medskip

The budget is shared across the entire contest rather than allocated
separately to each problem. Plan your attempts carefully to maximize the
total score.\par
\medskip

Current budget status:\par
- Counted actions used: \{ACTIONS\_USED\}\par
- Counted actions remaining: \{ACTIONS\_REMAINING\}\par
\medskip

Problems:\par
\medskip

===== Problem \{PROBLEM\_ID\_1\}: \{PROBLEM\_TITLE\_1\} =====\par
\{PROBLEM\_STATEMENT\_1\}\par
...\par
\medskip

[REMAINING PROBLEMS]\par
\end{promptbox}

\begin{promptbox}[Agentic Contest Prompt (continued)]
{green!50!black}{box:agentic_math_ar_prompt_submission}
\footnotesize\ttfamily\raggedright
\setlength{\parindent}{0pt}

Submission requirements:\par
\medskip

1. For each attempted problem, write the final submission in the following
format or location:\par
\medskip

\{PER\_PROBLEM\_SUBMISSION\_FORMAT\}\par
\medskip

2. Write the contest-level answer artifact to:\par
\medskip

\{ANSWER\_ARTIFACT\_PATH\}\par
\medskip

3. Every submitted answer must satisfy:\par
\medskip

\{ANSWER\_FORMAT\_REQUIREMENTS\}\par
\medskip

4. Problems without a valid submission are treated as unsubmitted and
receive zero credit.\par
\medskip

5. You may finish after submitting any subset of the problems. Solving all
\{NUM\_PROBLEMS\} problems is not required.\par
\medskip

6. When all desired submissions are ready, invoke the completion action by
itself. Do not combine it with another tool action.\par
\end{promptbox}

\begin{promptbox}[Agentic Contest Prompt (finalization)]
{green!50!black}{box:agentic_math_ar_prompt_finalization}
\footnotesize\ttfamily\raggedright
\setlength{\parindent}{0pt}

\textbf{[DYNAMIC BUDGET REMINDER]}\par
\medskip

Shared contest budget:\par
- Total counted actions: \{TOTAL\_BUDGET\}\par
- Counted actions used: \{ACTIONS\_USED\}\par
- Counted actions remaining: \{ACTIONS\_REMAINING\}\par
\medskip

This remaining budget must be shared across all \{NUM\_PROBLEMS\} problems.
Prioritize the actions most likely to improve the total contest score.\par
\bigskip

\textbf{[FINALIZATION PROMPT]}\par
\medskip

The normal problem-solving phase has ended. Finalize the contest now.\par
\medskip

Preserve all valid submissions already produced. You may submit fewer than
\{NUM\_PROBLEMS\} problems; unsubmitted problems receive zero credit.\par
\medskip

Do not continue exploratory work or begin a new problem. Ensure that:\par
\medskip

1. Each attempted problem has a valid final submission.\par
2. The contest-level artifact exists at \{ANSWER\_ARTIFACT\_PATH\}.\par
3. Every submitted answer follows \{ANSWER\_FORMAT\_REQUIREMENTS\}.\par
4. No incomplete or malformed answer is presented as a valid submission.\par
\medskip

After completing these bookkeeping steps, invoke the completion action by
itself.

\end{promptbox}

The with-difficulty variant inserts a difficulty line in the problem block; the
without-difficulty variant leaves this line empty.

\subsection{Judging Prompts}

Correctness is judged independently for each problem after the run. Mathematics
uses an offline model-based equivalence judge that compares the parsed student
answer against the reference answer while paying attention to mathematical
equivalence. We use DeepSeek V4 Flash~\citep{deepseekai2026deepseekv4} as the judge model. This judge prompt is never shown to the solver during the run.
Abstract reasoning uses the Reasoning Gym extractor and scorer, and coding uses executable tests through the offline verifier. 

\begin{promptbox}[Mathematics Equivalence-Judge Prompt]
{red!60!black}{box:math_judge_prompt}
\footnotesize\ttfamily\raggedright
Judge whether the student's final answer is mathematically equivalent to the
reference answer. The reference answer is assumed correct. Return a structured
report containing the student's final answer, a TRUE/FALSE equivalence judgment,
and a short justification.
\end{promptbox}

\subsection{Scheduler Prompt Variants}

The online scheduler directives in
Section~\ref{sec:recovering-allocation-gap} use the same agentic loop and change
only its runtime scheduler note. The paper labels differ from the implementation
scheme names: the Contest Reference (baseline) is implementation scheme~A,
paper Strategy~A is implementation scheme~B, and paper Strategy~B is
implementation scheme~C. Thus, the baseline has no scheduler guard,
Strategy~A adds coverage, and Strategy~B adds coverage plus verification. Full
implementation details are given in
Appendix~\ref{app:online_scheduler_directives}.

Because the three domains use the same scheduling logic, the box below presents
the shared directive once and then lists the domain-specific wording and
artifact names used by the runtime. It substitutes the values used in
Section~\ref{sec:recovering-allocation-gap}: six visible problems, one initial
probe per problem, and at most one additional check of an unchanged candidate.
The reported baseline/A/B comparison does not use oracle scheduling or the
implementation's stop-finalization scheme~D.
\begin{promptbox}[Online Scheduler Runtime Notes (Shared Template)]
{sharedbudget}{box:scheduler_prompt_variants}
\footnotesize\ttfamily\raggedright
\setlength{\parindent}{0pt}

\textbf{Paper baseline (implementation scheme A).}\par
No scheduler directive is added. The agent receives the common agentic,
native-tool, and budget prompts given above.\par
\bigskip

\textbf{Paper Strategy A (implementation scheme B: coverage).}\par
Coverage guard for this run:\par
- Before spending a second paid step on the same problem, give every visible
problem at least one paid focused probe.\par
- Use \texttt{focus\_problem i} before a paid command so the runtime can
attribute the step to one problem.\par
- If the guard blocks the current problem, shelve it and move to an unattempted
problem; finalization remains available.\par
\medskip

Runtime wording: Math and AR additionally state that six problems share one
paid compute/tool budget and use the headers ``Math runtime coverage guard''
and ``Suite runtime coverage guard,'' respectively. Code uses the header
``Code-domain online scheduler directive scheme B'' and the phrase ``paid
focused step(s)''; its baseline prompt already contains the focus/shelve
bookkeeping rules.\par
\bigskip

\textbf{Paper Strategy B (implementation scheme C: coverage + verification).}\par
Include the coverage guard above, followed by:\par
Cheap verification gate after coverage is complete:\par
- Continue a problem only when cheap non-oracle signals indicate that more
computation is useful.\par
- Once a candidate artifact is stable, allow at most one additional paid check
of that unchanged candidate.\par
- Changing or rewriting the candidate resets this allowance. The gate uses
candidate stability and format only, never oracle correctness.\par
\medskip

Runtime wording: Math and AR use the stable answer artifact
\texttt{/logs/artifacts/answer.txt} and the headers ``Math cheap verification
gate'' and ``Suite cheap verification gate.'' Code uses the header
``Code-domain online scheduler directive scheme C,'' treats
\texttt{solution\_X.cpp} as the candidate artifact, and specifies that the one
additional check is non-writing unless the solution is rewritten.\par
\end{promptbox}

\section{Response-Curve Oracle and Offline Knapsack Replay}
\label{app:oracle_knapsack}
To separate single-problem competence from cross-problem allocation, we
construct two offline replays from judged single-problem outcomes. Neither
replay reruns the model or generates new answers. The equal-allocation replay
asks how many observed successes fit under a uniform division of the contest
budget. The response-curve oracle instead asks how many fit under the best
allocation of the same budget. We make the dependence on evaluation setting
\(s\) explicit here; the main text suppresses this index for readability.

\paragraph{Budget grid.}
Fix a model \(m\), domain \(d\), and evaluation setting \(s\) (tool-free
reasoning or agentic reasoning). Let \(R^{\infty}_{m,d,s}\) denote the
unbudgeted resource use calibrated in
Appendix~\ref{app:budget_calibration}. The response curve is measured on a
fixed geometric grid of pressure levels
\[
\rho\in\Lambda=\{0.05,\,0.1,\,0.2,\,0.4,\,0.8\},
\]
with the corresponding resource level
\[
b^{\rho}_{m,d,s}
=
\bigl\lfloor \rho\,R^{\infty}_{m,d,s}\bigr\rfloor,
\]
together with a synthetic zero option \(b^{0}=0\), for which
\(q_p(0)=0\). This option represents leaving a problem unfunded and does
not invoke the model.

Each positive grid point is run \(K=5\) times per problem. Thus, each
problem contributes \(5|\Lambda|=25\) actual single-problem runs and six
replay options after adding the synthetic zero option. Runs are independent;
no run observes another run's output.

\paragraph{Response-curve observations.}
For a problem \(p\), positive grid level \(\ell\in\Lambda\), and independent
repeat \(k\in\{1,\ldots,K\}\), let
\(v_{p,\ell,k}\in\{0,1\}\) denote the judged outcome and let
\(C_{p,\ell,k}\) denote the realized consumption of the metered resource,
measured in the same unit as the corresponding contest budget.
A value of \(v_{p,\ell,k}=1\) means that the run is correct: in coding, the
submitted program is accepted by the executable verifier; in mathematics and
abstract reasoning, the extracted answer is judged correct by the
corresponding parser and judge.

The response curve records the empirical success rate at each positive level,
\[
q_p(\ell)
=
\frac{1}{K}\sum_{k=1}^{K} v_{p,\ell,k}
\in
\left\{
0,\tfrac{1}{5},\tfrac{2}{5},\tfrac{3}{5},\tfrac{4}{5},1
\right\},
\qquad \ell\in\Lambda,
\]
and we set \(q_p(0)=0\) for the synthetic zero level. These empirical success
rates provide the value terms used by the response-curve oracle below.

For the oracle, the cost assigned to level \(\ell\) is its nominal cap,
\[
c_p(\ell)=b^\ell_{m,d,s},
\]
rather than the realized consumption of any individual repeat. This makes the
cost of an oracle option fixed by its response-curve level and independent of
how much of the allowance a particular run happened to consume. The
equal-allocation replay instead operates at the attempt level and uses the
realized costs \(C_{p,\ell,k}\).

\paragraph{Observed-cost equal-share replay.}
For each problem \(p\), let
\[
\mathcal{S}_p
=
\left\{
(\ell,k)\in\Lambda\times\{1,\ldots,K\}:
v_{p,\ell,k}=1
\right\}
\]
denote the set of judged-correct positive-budget single-problem attempts.
Missing, malformed, unparsed, unjudged, and incorrect attempts are therefore
excluded.

We define the minimum observed successful cost of problem \(p\) as
\[
C_p^\star
=
\begin{cases}
\displaystyle
\min_{(\ell,k)\in\mathcal{S}_p} C_{p,\ell,k},
&
\mathcal{S}_p\neq\emptyset,
\\[6pt]
+\infty,
&
\mathcal{S}_p=\emptyset.
\end{cases}
\]

For a six-problem contest \(c=\{p_1,\ldots,p_6\}\) with total shared budget
\(B\), the replay assigns every problem the same cost threshold
\[
\tau(B)
=
\left\lfloor\frac{B}{6}\right\rfloor.
\]
Its suite-level score is
\[
\mathrm{Equal}_{m,d,s}(c,B)
=
\sum_{p\in c}
\mathbf{1}
\left[
C_p^\star\le \tau(B)
\right].
\]

Thus, each problem contributes one point if the collected single-problem runs
contain at least one judged-correct attempt whose realized resource cost does
not exceed one sixth of the matching contest budget. The criterion is based
on realized cost rather than nominal response-curve cap: an attempt launched
with a cap larger than \(\tau(B)\) may still qualify if it terminates
successfully within the threshold.

This replay is a post-hoc diagnostic rather than a literal rerun under six
independent hard caps of \(B/6\). It measures whether low-cost success was
observed in the collected single-problem runs, not whether the same success
would necessarily be reproduced under a newly imposed cap of that size. It
does not generate new model outputs, transfer unused shares across problems,
use difficulty labels, or adapt one problem's threshold based on outcomes
from other problems.

\paragraph{Response-curve oracle.}
The oracle reallocates the same total budget \(B\) across the six problems.
Each problem may be assigned at most one grid level, and a problem left
unassigned receives the zero level. Let \(z_{p,\ell}\in\{0,1\}\) indicate that
problem \(p\) is assigned level \(\ell\). The oracle solves
\[
\mathrm{Oracle}_{m,d,s}(c,B)
=
\max_{z}
\sum_{p\in c}
\sum_{\ell\in\Lambda\cup\{0\}}
q_{p}(\ell)\,z_{p,\ell}
\]
subject to
\[
\sum_{p\in c}
\sum_{\ell\in\Lambda\cup\{0\}}
b^{\ell}_{m,d,s}\,z_{p,\ell}
\le B,
\qquad
\sum_{\ell\in\Lambda\cup\{0\}} z_{p,\ell}=1
\ \ \forall p\in c .
\]
This is a multiple-choice knapsack: the oracle
chooses not only which problems to fund but at which level to fund them. The
formulation does not assume that \(q_{p}\) increases with the level, so the
oracle is free to fund a problem at a cheaper level whose observed success rate
is higher. Because each contest has six problems and each problem has six
options, we solve the program exactly by enumerating all \(6^{6}=46{,}656\)
assignments, which avoids budget-indexed dynamic programming when token
budgets are large.

When several assignments attain the same maximum, we select the one with the
smallest total cost. Any remaining ties are resolved by the fixed input order.
This rule affects which levels the oracle selects, but not the oracle score.

\paragraph{Aggregation and oracle gap.}
Let \(\mathcal{C}_d\) be the 50 contests in domain \(d\). Each contest is run
\(K=5\) times under the shared budget, and
\(\mathrm{Actual}_{m,d,s,\rho}(c)\in[0,6]\) denotes the mean judged score of
those runs. The reported contest score is
\[
\mathrm{Contest}_{m,d,s,\rho}
=
\frac{1}{|\mathcal{C}_d|}
\sum_{c\in\mathcal{C}_d}
\mathrm{Actual}_{m,d,s,\rho}(c),
\]
and the oracle score is
\[
\mathrm{Oracle}_{m,d,s,\rho}
=
\frac{1}{|\mathcal{C}_d|}
\sum_{c\in\mathcal{C}_d}
\mathrm{Oracle}_{m,d,s}
\!\left(c,R^\rho_{m,d,s}\right),
\]
where \(R^\rho_{m,d,s}\) is the calibrated shared budget at pressure level
\(\rho\). In tool-free reasoning,
\(R^\rho_{m,d,s}=B^\rho_{m,d}\) is an output-token budget. In the agentic
setting, \(R^\rho_{m,d,s}=A^\rho_{m,d}\) is a counted-action budget. Both
sides of the comparison therefore average over the same 50 contests and the
same number of repeats, and both are expected solved counts rather than
counts of individual runs.

The resource-rationality gap is
\[
\Delta_{\mathrm{RR}}
=
\mathrm{Oracle}_{m,d,s,\rho}
-
\mathrm{Contest}_{m,d,s,\rho}.
\]
The normalized gap reported in the main results is
\[
\mathrm{GapRatio}
=
\frac{
\mathrm{Oracle}_{m,d,s,\rho}
-
\mathrm{Contest}_{m,d,s,\rho}
}{
\mathrm{Oracle}_{m,d,s,\rho}
},
\]
computed only for cells with a nonzero oracle score.

\paragraph{Oracle-selected misses.}
Let \(z^\star(c,B)\) denote the deterministic optimal assignment returned by
the oracle after applying the tie-breaking rule above. The oracle-selected set
collects the problems it funds with a positive expected return,
\[
O(c,B)
=
\left\{
p\in c:
\textstyle\sum_{\ell}q_{p}(\ell)\,z^{\star}_{p,\ell}(c,B)>0
\right\}.
\]
Write \(q^{\star}_{p}\) for that selected success rate and
\(\mathrm{AC}_{m,d,s,\rho}(c,p)\in[0,1]\) for the actual contest success rate
of problem \(p\) over the \(K\) contest repeats. The recoverable mass on a
selected problem is
\[
\mu(c,p)
=
\max\!\left\{0,\;
q^{\star}_{p}-\mathrm{AC}_{m,d,s,\rho}(c,p)
\right\},
\]
and the selected-miss mass of a contest is \(\sum_{p\in O(c,B)}\mu(c,p)\).
This quantity reduces to the number of fully missed selected problems when all
rates are \(0\) or \(1\), and otherwise credits partial recovery. These
selected misses identify candidate recoverable allocation losses under the
replay assumptions. We aggregate them by difficulty tier, presented position,
and trajectory-level annotation.

\paragraph{Scope.}
The observed-cost equal-allocation replay is not a literal rerun of six
single-problem evaluations under newly imposed caps of \(B/6\). A model may
behave differently when rerun under a stricter nominal cap, even if an attempt
generated under a larger cap happened to terminate successfully after consuming
no more than \(B/6\). Equal therefore measures best-observed low-cost success
in the collected single-problem runs, rather than success that would
necessarily be reproduced under six independent hard caps of \(B/6\). As a
best-observed diagnostic, it also inherits finite-sample variation from the
collected attempts.

The response-curve oracle is optimal within the empirical replay defined by
the grid and the observed repeats. It is not a theoretical upper bound on the
model's contest performance, for two reasons. The replay can only fund a
problem at one of the grid levels, so it may miss allocations available on a
finer grid; and it inherits the sampling error of \(K=5\) repeats per grid
point. We therefore interpret the oracle gap as diagnostic allocation
headroom under observed single-problem evidence, not as performance that the
model is guaranteed to achieve in a real contest trajectory.
\section{Action Accounting and Tool Whitelist in the Agentic Setting}
\label{app:white_list}

Tool actions are classified according to whether they perform
problem-solving computation. The accounting rules separate paid computation
from commands that only inspect the environment, maintain runtime state, stage
files, or record final answers. The same rules apply to mathematics,
competitive programming, and abstract reasoning.

\paragraph{Action accounting unit.}
The agentic budget is enforced over parsed executable actions rather than model
turns. A single model turn may contain zero, one, or multiple commands. After
parsing, each command is classified independently as a counted action, a free
action, a blocked action, or a protocol error. Each executed counted action has
unit cost one. A counted command consumes one budget unit once it is accepted
for execution, even if it later exits with a nonzero status, produces an error,
or times out. Commands blocked before execution do not increase the executed
counted-action counter. Some runtimes also record stricter diagnostic counters
for blocked attempts or malformed turns. These counters are not the official
resource unit reported in the main results.

\paragraph{Active budget policy.}
All three domains---mathematics, competitive programming, and abstract
reasoning---use the \texttt{compute\_tools} policy. Under this policy, a command
is counted if it invokes computation used for solving or verification. Counted
operations include Python and other interpreters, calculators, compilers,
locally built executables, local tests, solver-like programs, shell arithmetic,
and text- or data-processing commands. Representative examples include
\texttt{python}, \texttt{awk}, \texttt{bc}, \texttt{node}, \texttt{g++},
\texttt{java}, \texttt{grep}, \texttt{rg}, \texttt{sed}, and \texttt{sort}.
A command that invokes any such operation consumes one action, regardless of
how many compute operations it contains.

Commands that do not perform computation are not counted. These include
bookkeeping helpers, simple environment inspection, passive file operations,
pure source-file staging, and final-artifact writes. For example, writing a
program to a file is free, whereas compiling, executing, or testing that
program is counted. Similarly, recording an answer is free, whereas computation
used to derive or verify it is counted. Non-compute commands are retained in
the trajectory and logged separately.

Table~\ref{tab:tool_whitelist} summarizes the command classes, their budget
status, and their role in the runtime.

\begin{table*}[t]
\centering
\small
\setlength{\tabcolsep}{4pt}
\begin{tabular}{
    p{0.21\linewidth}
    p{0.27\linewidth}
    p{0.16\linewidth}
    p{0.25\linewidth}
}
\toprule
Action type & Examples & Budget status & Role \\
\midrule

Compute command &
\texttt{python}, \texttt{python3}, \texttt{perl}, \texttt{awk},
\texttt{bc}, \texttt{node}, \texttt{Rscript}, \texttt{julia} &
Counted &
Performs computation used for solving or verification. \\

Compilation or execution &
\texttt{gcc}, \texttt{g++}, \texttt{javac}, \texttt{java}, locally built
executables, local tests &
Counted &
Compiles, executes, or tests a candidate solution. \\

Text or data processing &
\texttt{grep}, \texttt{rg}, \texttt{sed}, \texttt{wc}, \texttt{tr},
\texttt{sort}, \texttt{uniq}, \texttt{cut}, \texttt{paste}, \texttt{seq} &
Counted &
Processes task-relevant text or data during solving. \\

Shell computation &
Shell arithmetic, \texttt{expr}, \texttt{let}, or arithmetic loops &
Counted &
Performs computation directly through the shell. \\

Passive file operation &
Direct file reads, copies, edits, or pure staging of source and scratch files &
Free non-compute action &
Inspects or records task state without executing computation. \\

Final artifact write &
Writing \texttt{/app/solution\_X.cpp},
\texttt{/logs/artifacts/answer.txt}, or an answer slot &
Free finalization &
Records an existing solution or answer without returning correctness feedback. \\

Focus bookkeeping &
\texttt{focus\_problem \textless id\textgreater} &
Free &
Sets the active problem for per-problem attribution. \\

Shelve bookkeeping &
\texttt{shelve\_problem \textless id\textgreater\ \textless note\textgreater} &
Free &
Records that the agent is switching away from a problem. \\

Status query &
\texttt{contest\_status} &
Free &
Returns sanitized focus and artifact status without verdict feedback. \\

Problem-scoped answer helper &
\texttt{submit\_answer \textless id\textgreater\ \textless answer\textgreater} &
Free when provided &
Updates one answer slot without revealing whether the answer is correct. \\

Task completion signal &
\texttt{mark\_task\_complete} or \texttt{task\_complete} &
Free when provided &
Ends the episode; the verifier later grades the existing artifacts. \\

Simple environment inspection &
\texttt{pwd}; simple \texttt{ls}; \texttt{which}/\texttt{type};
restricted \texttt{cat} &
Free in whitelisted forms &
Inspects runtime state without performing problem-solving computation. \\

Mixed shell command &
A file operation combined with compilation, execution, arithmetic, or
text processing &
Counted &
Consumes one action because the command contains a compute operation. \\

Blocked command &
Paid computation after budget exhaustion, paid computation without active
focus, cross-problem access, or a live-judge request &
Not executed &
Enforces the shared budget, problem scope, and no-feedback constraints. \\

Protocol error &
Malformed JSON, malformed native tool call, or empty command turn &
Not counted as an executed action &
Logged as a protocol failure or overhead. \\

\bottomrule
\end{tabular}
\caption{Action-accounting rules under the shared
\texttt{compute\_tools} policy. The same policy is used for mathematics,
competitive programming, and abstract reasoning. Counted actions perform
problem-solving computation. Free actions inspect or record state without
revealing correctness.}
\label{tab:tool_whitelist}
\end{table*}

\paragraph{Budget exhaustion and answer eligibility.}
Before executing a compute command, the runtime checks the remaining shared
budget. If budget remains, the command consumes one unit and is executed. Its
cost is not refunded if it fails, exits with an error, or times out. Once the
budget is exhausted, subsequent compute commands are blocked before execution
and logged as blocked commands. They do not increase the executed
counted-action counter and do not invalidate artifacts produced earlier in the
episode.

Non-compute bookkeeping and finalization actions remain available after compute
budget exhaustion. The agent may therefore record solutions already obtained,
write the corresponding final artifacts, and invoke the completion action. It
may not compile, execute, test, calculate, search, or otherwise process data to
improve those solutions. Only answers present in the designated final artifacts
are eligible for credit.

\paragraph{Per-problem attribution.}
In problem-scoped contest runs, every counted action must be associated with an
active problem. The agent declares the active problem with
\texttt{focus\_problem}; \texttt{shelve\_problem} records a switch away from
that problem. These commands are free because they update attribution state
only. If a paid command is issued without an active focus, or if it accesses
another problem's scratch area or multiple problem slots, the runtime blocks
the command rather than leaving it unattributed. This rule controls action
accounting and scope. It does not constrain natural-language planning within
the model's messages.

\paragraph{No interactive correctness feedback.}
Official correctness feedback is unavailable during a run. Competitive
programming agents may compile programs and run local tests, but these actions
consume budget. Live-judge and submit-style commands are blocked. Mathematics
and abstract-reasoning agents may use permitted local computation under the
same \texttt{compute\_tools} policy. In every domain, correctness is determined
only after the episode from the designated final artifacts. Bookkeeping,
status, and finalization helpers do not provide verification feedback.

\paragraph{Logging.}
Each trajectory records the configured budget, used budget, remaining budget,
free bookkeeping and environment steps, non-compute audit steps, blocked
commands, active focus, per-problem attribution when available, final
artifacts, and final verifier outputs. Field names differ slightly across
domains, but the accounting policy is shared and correctness is always computed
separately by the final domain-specific verifier.
\section{Budget Calibration}
\label{app:budget_calibration}

Table~\ref{tab:budget_calibration} records the resource caps used in the
formal without-difficulty contest runs. Pure natural-language reasoning uses
output-token caps \(B^{\rho}_{m,d}\). Agentic reasoning uses executed
counted-action caps \(A^{\rho}_{m,d}\). The values come directly from the
frozen six-model cell inventory. They are not reconstructed from an assumed
unconstrained baseline. Appendix~\ref{app:white_list} defines action
accounting.

\begin{table}[!htbp]
\centering
\small
\setlength{\tabcolsep}{8pt}
\renewcommand{\arraystretch}{1.06}
\resizebox{\columnwidth}{!}{%
\begin{tabular}{llrrrr}
\toprule
& & \multicolumn{2}{c}{Tool-free reasoning} & \multicolumn{2}{c}{Agentic reasoning} \\
\cmidrule(lr){3-4}\cmidrule(l){5-6}
Model & Domain & $B^{0.2}$ & $B^{0.8}$ & $A^{0.2}$ & $A^{0.8}$ \\
\midrule
DeepSeek-V4-Pro & Math & 19161 & 76642 & 2 & 8 \\
DeepSeek-V4-Pro & Code & 16800 & 67200 & 7 & 28 \\
DeepSeek-V4-Pro & AR & 15500 & 62000 & 2 & 7 \\
\midrule
Qwen3.7-Max & Math & 8317 & 33265 & 2 & 10 \\
Qwen3.7-Max & Code & 25500 & 101989 & 3 & 12 \\
Qwen3.7-Max & AR & 9303 & 37212 & 2 & 8 \\
\midrule
GLM-5.2 & Math & 1412 & 5647 & 3 & 13 \\
GLM-5.2 & Code & 834 & 3336 & 7 & 28 \\
GLM-5.2 & AR & 10992 & 43966 & 2 & 8 \\
\midrule
Hy-3 & Math & 6141 & 24565 & 2 & 7 \\
Hy-3 & Code & 12243 & 48970 & 8 & 32 \\
Hy-3 & AR & 6345 & 25379 & 2 & 8 \\
\midrule
GPT-5.5 & Math & 229 & 913 & 2 & 7 \\
GPT-5.5 & Code & 287 & 1147 & 2 & 10 \\
GPT-5.5 & AR & 36 & 141 & 2 & 6 \\
\midrule
Claude-Opus-4.8 & Math & 472 & 1888 & 1 & 6 \\
Claude-Opus-4.8 & Code & 1054 & 4216 & 4 & 16 \\
Claude-Opus-4.8 & AR & 480 & 1919 & 1 & 6 \\
\bottomrule
\end{tabular}%
}
\caption{Formal shared-budget caps for the six-model evaluation. Pure-NL budgets are output-token caps; agentic budgets are counted-action caps. Every value is read from the frozen without-difficulty cell inventory.}
\label{tab:budget_calibration}
\end{table}

\section{Detailed Shared-Budget Contest Results}
\label{app:full_result_breakdown}

\begin{table*}[htbp]
\centering
\scriptsize
\setlength{\tabcolsep}{3.8pt}
\renewcommand{\arraystretch}{1.12}
\caption{
\textbf{Main results on \bench{} with domains as columns.}
For each domain, Contest and Oracle report the average number of correct answers per six-problem suite.
Gap Ratio denotes $\Delta_{\mathrm{RR}}/\mathrm{Oracle}$, where $\Delta_{\mathrm{RR}}=\mathrm{Oracle}-\mathrm{Contest}$.
Lower Gap Ratio is better. AR denotes abstract reasoning.
}
\label{tab:full-results-domain-columns}
\resizebox{\textwidth}{!}{%
\begin{tabular}{@{}lll *{9}{c}@{}}
\toprule
\textbf{Setting}
& \textbf{Model}
& \textbf{Regime}
& \multicolumn{3}{c}{\textbf{Math}}
& \multicolumn{3}{c}{\textbf{Code}}
& \multicolumn{3}{c}{\textbf{AR}} \\
\cmidrule(lr){4-6}
\cmidrule(lr){7-9}
\cmidrule(l){10-12}
& &
& \textbf{Contest} & \textbf{Oracle} & \textbf{Gap Ratio}
& \textbf{Contest} & \textbf{Oracle} & \textbf{Gap Ratio}
& \textbf{Contest} & \textbf{Oracle} & \textbf{Gap Ratio} \\
\midrule

\multirow{16}{*}{\makecell[l]{Tool-free}}
& \multirow{2}{*}{DeepSeek-Chat}
& $\rho{=}0.2$ & 2.42 & 3.24 & 25.31\% & 0.86 & 2.02 & 57.43\% & 1.80 & 3.00 & 40.00\% \\
& & $\rho{=}0.8$ & 3.40 & 4.06 & 16.26\% & 1.68 & 2.24 & 25.00\% & 2.80 & 3.88 & 27.84\% \\
\cmidrule(lr){2-12}

& \multirow{2}{*}{DeepSeek-Reasoner}
& $\rho{=}0.2$ & 2.42 & 3.68 & 34.24\% & 1.38 & 1.88 & 26.60\% & 2.14 & 3.64 & 41.21\% \\
& & $\rho{=}0.8$ & 3.70 & 4.98 & 25.70\% & 2.44 & 3.34 & 26.95\% & 2.98 & 4.32 & 31.02\% \\
\cmidrule(lr){2-12}

& \multirow{2}{*}{DeepSeek-V4-Pro}
& $\rho{=}0.2$ & 2.34 & 3.94 & 40.61\% & 1.38 & 3.12 & 55.77\% & 2.24 & 4.02 & 44.28\% \\
& & $\rho{=}0.8$ & 3.94 & 5.08 & 22.44\% & 3.06 & 4.62 & 33.77\% & 2.90 & 4.64 & 37.50\% \\
\cmidrule(lr){2-12}

& \multirow{2}{*}{Qwen3.7-Max}
& $\rho{=}0.2$ & 3.54 & 3.70 & 4.32\% & 1.42 & 2.94 & 51.70\% & 2.76 & 4.06 & 32.02\% \\
& & $\rho{=}0.8$ & 4.32 & 5.06 & 14.62\% & 3.44 & 5.24 & 34.35\% & 3.82 & 4.70 & 18.72\% \\
\cmidrule(lr){2-12}

& \multirow{2}{*}{GLM-5.2}
& $\rho{=}0.2$ & 2.32 & 3.08 & 24.68\% & 0.68 & 1.88 & 63.83\% & 1.14 & 3.51 & 67.52\% \\
& & $\rho{=}0.8$ & 2.72 & 3.72 & 26.88\% & 1.72 & 2.08 & 17.31\% & 1.70 & 4.19 & 59.43\% \\
\cmidrule(lr){2-12}

& \multirow{2}{*}{Hy-3}
& $\rho{=}0.2$ & 2.40 & 3.36 & 28.57\% & 1.20 & 2.28 & 47.37\% & 1.56 & 3.11 & 49.84\% \\
& & $\rho{=}0.8$ & 3.72 & 4.68 & 20.51\% & 2.96 & 3.88 & 23.71\% & 3.16 & 4.16 & 24.04\% \\
\cmidrule(lr){2-12}

& \multirow{2}{*}{GPT-5.5}
& $\rho{=}0.2$ & 0.72 & 1.28 & 43.75\% & 0.68 & 1.28 & 46.88\% & 0.34 & 1.94 & 82.47\% \\
& & $\rho{=}0.8$ & 2.52 & 2.80 & 10.00\% & 2.60 & 3.12 & 16.67\% & 1.58 & 2.75 & 42.55\% \\
\cmidrule(lr){2-12}

& \multirow{2}{*}{Claude-Opus-4.8}
& $\rho{=}0.2$ & 3.78 & 4.30 & 12.09\% & 0.60 & 1.88 & 68.09\% & 1.84 & 3.11 & 40.82\% \\
& & $\rho{=}0.8$ & 4.18 & 4.96 & 15.73\% & 2.68 & 3.48 & 22.99\% & 2.52 & 4.15 & 39.21\% \\

\midrule

\multirow{16}{*}{Agentic}
& \multirow{2}{*}{DeepSeek-Chat}
& $\rho{=}0.2$ & 4.24 & 5.08 & 16.54\% & 2.44 & 3.98 & 38.69\% & 4.00 & 5.08 & 21.26\% \\
& & $\rho{=}0.8$ & 4.68 & 5.26 & 11.03\% & 3.86 & 5.10 & 24.31\% & 4.64 & 5.26 & 11.79\% \\
\cmidrule(lr){2-12}

& \multirow{2}{*}{DeepSeek-Reasoner}
& $\rho{=}0.2$ & 4.38 & 5.40 & 18.89\% & 3.08 & 4.00 & 23.00\% & 4.42 & 5.26 & 15.97\% \\
& & $\rho{=}0.8$ & 4.72 & 5.44 & 13.24\% & 4.10 & 4.98 & 17.67\% & 4.74 & 5.28 & 10.23\% \\
\cmidrule(lr){2-12}

& \multirow{2}{*}{DeepSeek-V4-Pro}
& $\rho{=}0.2$ & 4.48 & 5.28 & 15.15\% & 2.60 & 4.90 & 46.94\% & 4.36 & 5.06 & 13.83\% \\
& & $\rho{=}0.8$ & 4.48 & 5.40 & 17.04\% & 4.10 & 5.53 & 25.86\% & 4.12 & 5.14 & 19.84\% \\
\cmidrule(lr){2-12}

& \multirow{2}{*}{Qwen3.7-Max}
& $\rho{=}0.2$ & 4.64 & 5.14 & 9.73\% & 2.78 & 2.96 & 6.08\% & 4.62 & 4.96 & 6.85\% \\
& & $\rho{=}0.8$ & 4.96 & 5.34 & 7.12\% & 4.72 & 5.46 & 13.55\% & 4.80 & 5.14 & 6.61\% \\
\cmidrule(lr){2-12}

& \multirow{2}{*}{GLM-5.2}
& $\rho{=}0.2$ & 4.48 & 4.58 & 2.18\% & 2.02 & 3.90 & 48.21\% & 3.78 & 4.52 & 16.37\% \\
& & $\rho{=}0.8$ & 4.16 & 5.12 & 18.75\% & 4.06 & 5.06 & 19.76\% & 4.06 & 5.04 & 19.44\% \\
\cmidrule(lr){2-12}

& \multirow{2}{*}{Hy-3}
& $\rho{=}0.2$ & 1.80 & 3.38 & 46.75\% & 1.00 & 3.00 & 66.67\% & 1.22 & 3.14 & 61.15\% \\
& & $\rho{=}0.8$ & 1.56 & 4.16 & 62.50\% & 1.38 & 3.56 & 61.24\% & 0.80 & 4.32 & 81.48\% \\
\cmidrule(lr){2-12}

& \multirow{2}{*}{GPT-5.5}
& $\rho{=}0.2$ & 3.64 & 4.96 & 26.61\% & 1.92 & 2.00 & 4.00\% & 1.80 & 4.92 & 63.41\% \\
& & $\rho{=}0.8$ & 4.40 & 5.32 & 17.29\% & 4.80 & 5.53 & 13.25\% & 4.72 & 5.24 & 9.92\% \\
\cmidrule(lr){2-12}

& \multirow{2}{*}{Claude-Opus-4.8}
& $\rho{=}0.2$ & 4.46 & 4.46 & 0.00\% & 2.24 & 3.56 & 37.08\% & 3.77 & 3.88 & 2.84\% \\
& & $\rho{=}0.8$ & 4.85 & 5.02 & 3.39\% & 4.52 & 5.08 & 11.02\% & 5.12 & 5.34 & 4.12\% \\
\bottomrule
\end{tabular}%
}
\end{table*}
Tables~\ref{tab:pure_nl_by_difficulty} and
\ref{tab:agentic_by_difficulty} provide the difficulty-level accuracy
breakdown for the formal shared-budget contest runs. Each model--budget
condition contains the same 50 six-problem suites, totaling 150 easy,
100 medium, and 50 hard problem instances, and each suite is run five times.
Entries are percentages over judged runs: the E, M, and H columns report
accuracy within each difficulty tier over 750, 500, and 250 runs respectively,
and the All column reports accuracy over all 1500 runs.
\begin{table*}[t]
\centering
\scriptsize
\setlength{\tabcolsep}{3pt}
\renewcommand{\arraystretch}{0.95}
\captionsetup{skip=2pt}
\caption{Tool-free shared-budget contest accuracy by budget
pressure, domain, and difficulty tier. Entries are percentages over the
problems evaluated in each condition.}
\label{tab:pure_nl_by_difficulty}
\resizebox{\textwidth}{!}{%
\begin{tabular}{@{}ll *{12}{c}@{}}
\toprule
Model & \(\rho\)
& \multicolumn{4}{c}{Math}
& \multicolumn{4}{c}{Code}
& \multicolumn{4}{c}{AR} \\
\cmidrule(lr){3-6}
\cmidrule(lr){7-10}
\cmidrule(l){11-14}
& & E & M & H & All & E & M & H & All & E & M & H & All \\
\midrule

\multirow[c]{2}{*}{DeepSeek-Chat}
& 0.2
& 60.7 & 22.0 & 16.0 & 40.3
& 28.0 & 1.0 & 0.0 & 14.3
& 42.7 & 22.0 & 8.0 & 30.0 \\
& 0.8
& 76.0 & 43.0 & 26.0 & 56.7
& 54.7 & 2.0 & 0.0 & 28.0
& 58.0 & 44.0 & 18.0 & 46.7 \\

\midrule
\multirow[c]{2}{*}{DeepSeek-Reasoner}
& 0.2
& 56.7 & 30.0 & 12.0 & 40.3
& 41.3 & 7.0 & 0.0 & 23.0
& 49.3 & 27.0 & 12.0 & 35.7 \\
& 0.8
& 73.3 & 57.0 & 36.0 & 61.7
& 65.3 & 22.0 & 4.0 & 40.7
& 66.7 & 42.0 & 14.0 & 49.7 \\

\midrule
\multirow[c]{2}{*}{DeepSeek-V4-Pro}
& 0.2
& 58.7 & 26.0 & 6.0 & 39.0
& 36.0 & 14.0 & 2.0 & 23.0
& 50.7 & 31.0 & 10.0 & 37.3 \\
& 0.8
& 77.3 & 60.0 & 42.0 & 65.7
& 70.7 & 44.0 & 6.0 & 51.0
& 61.3 & 47.0 & 12.0 & 48.3 \\

\midrule
\multirow[c]{2}{*}{Qwen3.7-Max}
& 0.2
& 78.7 & 51.0 & 16.0 & 59.0
& 33.3 & 20.0 & 2.0 & 23.7
& 63.3 & 34.0 & 18.0 & 46.0 \\
& 0.8
& 85.3 & 64.0 & 48.0 & 72.0
& 70.0 & 57.0 & 20.0 & 57.3
& 84.7 & 52.0 & 24.0 & 63.7 \\

\midrule
\multirow[c]{2}{*}{GLM-5.2}
& 0.2
& 59.3 & 20.0 & 14.0 & 38.7
& 22.7 & 0.0 & 0.0 & 11.3
& 34.0 & 6.0 & 0.0 & 19.0 \\
& 0.8
& 68.7 & 25.0 & 16.0 & 45.3
& 56.0 & 2.0 & 0.0 & 28.7
& 46.7 & 7.0 & 16.0 & 28.3 \\

\midrule
\multirow[c]{2}{*}{Hy-3}
& 0.2
& 62.7 & 18.0 & 16.0 & 40.0
& 32.0 & 12.0 & 0.0 & 20.0
& 40.0 & 16.0 & 4.0 & 26.0 \\
& 0.8
& 84.0 & 40.0 & 40.0 & 62.0
& 73.3 & 36.0 & 4.0 & 49.3
& 65.3 & 46.0 & 28.0 & 52.7 \\

\midrule
\multirow[c]{2}{*}{GPT-5.5}
& 0.2
& 22.7 & 2.0 & 0.0 & 12.0
& 21.3 & 2.0 & 0.0 & 11.3
& 8.7 & 4.0 & 0.0 & 5.7 \\
& 0.8
& 64.0 & 26.0 & 8.0 & 42.0
& 76.0 & 16.0 & 0.0 & 43.3
& 41.3 & 13.0 & 8.0 & 26.3 \\

\midrule
\multirow[c]{2}{*}{Claude-Opus-4.8}
& 0.2
& 70.7 & 53.0 & 60.0 & 63.0
& 18.7 & 2.0 & 0.0 & 10.0
& 47.3 & 12.0 & 18.0 & 30.7 \\
& 0.8
& 80.7 & 59.0 & 58.0 & 69.7
& 62.7 & 36.0 & 8.0 & 44.7
& 60.0 & 30.0 & 12.0 & 42.0 \\

\bottomrule
\end{tabular}%
}
\end{table*}

\begin{table*}[t]
\centering
\scriptsize
\setlength{\tabcolsep}{3pt}
\renewcommand{\arraystretch}{0.95}
\captionsetup{skip=2pt}
\caption{Agentic shared-budget contest accuracy by budget pressure, domain,
and difficulty tier. Entries are percentages over the problems evaluated in
each condition.}
\label{tab:agentic_by_difficulty}
\resizebox{\textwidth}{!}{%
\begin{tabular}{@{}ll *{12}{c}@{}}
\toprule
Model & \(\rho\)
& \multicolumn{4}{c}{Math}
& \multicolumn{4}{c}{Code}
& \multicolumn{4}{c}{AR} \\
\cmidrule(lr){3-6}
\cmidrule(lr){7-10}
\cmidrule(l){11-14}
& & E & M & H & All & E & M & H & All & E & M & H & All \\
\midrule

\multirow[c]{2}{*}{DeepSeek-Chat}
& 0.2
& 85.3 & 62.0 & 44.0 & 70.7
& 75.3 & 9.0 & 0.0 & 40.7
& 76.7 & 60.0 & 50.0 & 66.7 \\
& 0.8
& 89.3 & 71.0 & 58.0 & 78.0
& 93.3 & 49.0 & 8.0 & 64.3
& 83.3 & 72.0 & 70.0 & 77.3 \\

\midrule
\multirow[c]{2}{*}{DeepSeek-Reasoner}
& 0.2
& 84.7 & 69.0 & 46.0 & 73.0
& 87.3 & 21.0 & 4.0 & 51.3
& 82.7 & 69.0 & 56.0 & 73.7 \\
& 0.8
& 90.0 & 70.0 & 62.0 & 78.7
& 96.7 & 57.0 & 6.0 & 68.3
& 82.7 & 75.0 & 76.0 & 79.0 \\

\midrule
\multirow[c]{2}{*}{DeepSeek-V4-Pro}
& 0.2
& 91.3 & 66.0 & 42.0 & 74.7
& 70.0 & 24.0 & 2.0 & 43.3
& 81.3 & 64.0 & 64.0 & 72.7 \\
& 0.8
& 92.7 & 66.0 & 38.0 & 74.7
& 89.3 & 65.0 & 12.0 & 68.3
& 78.0 & 61.0 & 56.0 & 68.7 \\

\midrule
\multirow[c]{2}{*}{Qwen3.7-Max}
& 0.2
& 86.7 & 74.0 & 56.0 & 77.3
& 86.0 & 9.0 & 2.0 & 46.3
& 83.3 & 68.0 & 76.0 & 77.0 \\
& 0.8
& 90.7 & 79.0 & 66.0 & 82.7
& 98.7 & 76.0 & 24.0 & 78.7
& 87.3 & 73.0 & 72.0 & 80.0 \\

\midrule
\multirow[c]{2}{*}{GLM-5.2}
& 0.2
& 88.0 & 70.0 & 44.0 & 74.7
& 54.0 & 19.0 & 2.0 & 33.7
& 68.7 & 61.0 & 50.0 & 63.0 \\
& 0.8
& 84.7 & 60.0 & 42.0 & 69.3
& 85.3 & 66.0 & 18.0 & 67.7
& 80.0 & 57.0 & 52.0 & 67.7 \\

\midrule
\multirow[c]{2}{*}{Hy-3}
& 0.2
& 41.3 & 20.0 & 16.0 & 30.0
& 25.3 & 12.0 & 0.0 & 16.7
& 32.7 & 9.0 & 6.0 & 20.3 \\
& 0.8
& 32.7 & 22.0 & 14.0 & 26.0
& 42.0 & 6.0 & 0.0 & 23.0
& 18.7 & 9.0 & 6.0 & 13.3 \\

\midrule
\multirow[c]{2}{*}{GPT-5.5}
& 0.2
& 68.0 & 54.0 & 52.0 & 60.7
& 57.3 & 10.0 & 0.0 & 32.0
& 29.3 & 32.0 & 28.0 & 30.0 \\
& 0.8
& 82.7 & 68.0 & 56.0 & 73.3
& 96.0 & 82.0 & 28.0 & 80.0
& 76.0 & 78.0 & 88.0 & 78.7 \\

\midrule
\multirow[c]{2}{*}{Claude-Opus-4.8}
& 0.2
& 89.7 & 57.8 & 61.6 & 74.4
& 66.7 & 12.0 & 0.0 & 37.3
& 76.9 & 42.2 & 61.6 & 62.8 \\
& 0.8
& 89.7 & 69.2 & 77.2 & 80.8
& 90.7 & 78.0 & 24.0 & 75.3
& 86.1 & 83.2 & 87.6 & 85.4 \\

\bottomrule
\end{tabular}%
}
\end{table*}
\paragraph{Discriminant validity of Gap Ratio.}
Table~\ref{tab:eci_discriminant} reports the per-cell Spearman correlation
between model ECI and \bench{} Gap Ratio, with exact permutation \(p\)-values,
supporting the construct-validity analysis in
Section~\ref{subsec:effectiveness}. Capability ordering explains little of the
allocation-gap ordering: no cell reaches \(p<0.05\), the mean Spearman
correlation is \(-0.29\) (\(p=0.37\)), and the pooled within-cell \(R^2\) is
\(0.01\) (\(p=0.70\)).
\begin{table}[t]
\centering\small
\caption{Discriminant validity: Spearman correlation between model ECI and \bench{} Gap Ratio in each evaluation cell, with exact permutation $p$-values ($n{=}5$ models, $120$ permutations). Capability ordering explains little of the allocation-gap ordering. Global: mean Spearman $=-0.29$ ($p=0.37$); pooled within-cell $R^2=0.01$ ($p=0.70$).}
\label{tab:eci_discriminant}
\resizebox{\columnwidth}{!}{%
\begin{tabular}{@{}llccc@{}}
\toprule
Setting & Domain & $\rho$ & Spearman & Perm.\ $p$ \\
\midrule
Tool-Free & Math & 0.2 & +0.10 & 0.95 \\
Tool-Free & Code & 0.2 & -0.30 & 0.68 \\
Tool-Free & AR & 0.2 & +0.20 & 0.78 \\
Tool-Free & Math & 0.8 & -0.80 & 0.13 \\
Tool-Free & Code & 0.8 & -0.50 & 0.45 \\
Tool-Free & AR & 0.8 & +0.20 & 0.78 \\
Agentic & Math & 0.2 & +0.10 & 0.95 \\
Agentic & Code & 0.2 & -0.80 & 0.13 \\
Agentic & AR & 0.2 & +0.10 & 0.95 \\
Agentic & Math & 0.8 & -0.20 & 0.78 \\
Agentic & Code & 0.8 & -0.90 & 0.08 \\
Agentic & AR & 0.8 & -0.70 & 0.23 \\
\bottomrule
\end{tabular}}
\end{table}

\section{Model Interfaces and Experimental Configuration}
\label{app:model_card_config}

\paragraph{Evaluation interfaces.}
The benchmark has two evaluation interfaces. Tool-free reasoning is text-only: the model receives either one problem or a six-problem
contest prompt and returns final answers without shell access, code execution,
or interactive feedback. Agentic reasoning runs in the Harbor/Terminus-2 shell
environment under the action-accounting rules in Appendix~\ref{app:white_list}.

\paragraph{Model-call configuration.}
Table~\ref{tab:api_decoding_config} reports the model-call parameters used
for each model family. We use the sampling configurations recommended by the
model providers, including provider defaults where no task-specific setting is
given. We do not tune these parameters on \bench{}. For GPT-5.5 and
Claude-Opus-4.8, we disable thinking because their APIs do not expose the
corresponding chain-of-thought content in our evaluation interfaces. Each
model is evaluated with five independent runs per problem or contest instance.
For agentic evaluation, these are five complete trajectories rather than five
completions returned by a single model call. Budget-specific output limits are
reported in Appendix~\ref{app:budget_calibration}.

\begin{table*}[!t]
\centering
\small
\setlength{\tabcolsep}{4pt}
\begin{tabular}{
    p{0.20\textwidth}
    p{0.13\textwidth}
    p{0.13\textwidth}
    p{0.36\textwidth}
    p{0.08\textwidth}
}
\toprule
Model
& Temperature
& Top-p
& Reasoning / thinking configuration
& Independent runs \\
\midrule

DeepSeek-Chat
& \(0.0\)
& \(1.0\)
& Non-thinking chat completion.
& \(5\) \\

DeepSeek-Reasoner
& Omitted
& Omitted
& Native reasoning mode; sampling parameters are unsupported and have no
effect.
& \(5\) \\

DeepSeek-V4-Pro
& Omitted
& Omitted
& Thinking enabled with
\texttt{reasoning\_effort=high}.
& \(5\) \\

Qwen3.7-Max
& \(0.2\)
& \(0.8\)
& Thinking enabled for reasoning runs and disabled for answer-only
finalization.
& \(5\) \\

GLM-5.2
& \(1.0\)
& \(0.95\)
& Thinking enabled with high effort for reasoning and agentic tasks.
& \(5\) \\

Hy-3
& \(0.9\)
& \(1.0\)
& \texttt{reasoning\_effort=high} for mathematics, coding, and abstract
reasoning.
& \(5\) \\

GPT-5.5
& Omitted
& Omitted
& Thinking disabled with
\texttt{reasoning\_effort=none}, because chain-of-thought content is not
exposed through the evaluation interface.
& \(5\) \\

Claude-Opus-4.8
& Omitted
& Omitted
& Thinking disabled because chain-of-thought content is not exposed through
the evaluation interface.
& \(5\) \\

\bottomrule
\end{tabular}
\caption{Model-call parameters used in the evaluation. Sampling parameters
follow official provider recommendations or provider defaults and are not
tuned on \bench{}. ``Omitted'' indicates that a parameter is unsupported or
not applicable. GPT-5.5 and Claude-Opus-4.8 are evaluated without thinking
because their chain-of-thought content is not visible through our evaluation
interfaces. Each reported configuration uses five independent runs per
instance.}
\label{tab:api_decoding_config}
\end{table*}

\paragraph{Agentic sandbox.}
For agentic runs, Table~\ref{tab:agentic_sandbox_config} records the sandbox
limits and logging surface. The table does not redefine action accounting or
answer grading; those are specified in Appendix~\ref{app:white_list} and
Appendix~\ref{app:answer_judging}.

\begin{table*}[!t]
\centering
\small
\setlength{\tabcolsep}{4pt}
\begin{tabular}{p{0.19\textwidth}p{0.35\textwidth}p{0.34\textwidth}}
\toprule
Configuration item & Coding & Mathematics / abstract reasoning \\
\midrule
Runtime container
& Docker Harbor task with Terminus-2 shell runtime.
& Docker Harbor job with Terminus-2 shell runtime. \\

Timeouts
& Agent timeout \(7200\)s; build timeout \(600\)s.
& Agent and verifier timeout \(7200\)s. \\

Sandbox resources
& Memory \(2048\) MB; storage \(10240\) MB.
& \(1\) CPU; memory \(2048\) MB; storage \(10240\) MB. \\

Tool environment
& Local shell, file editing, C++ compilation, local execution, and local tests under the \texttt{compute\_tools} policy.
& Local shell plus computation-oriented tools under the \texttt{compute\_tools} policy. \\

Trajectory logging
& Model messages, tool calls, shell commands, outputs, exposed reasoning content, budget usage, blocked commands, and final artifacts.
& Model messages, tool calls, shell commands, outputs, exposed reasoning content, budget usage, blocked commands, and final artifacts. \\
\bottomrule
\end{tabular}
\caption{Agentic sandbox and logging configuration. The table reports runtime
constraints material to the benchmark while omitting credentials, private server
identifiers, and deployment-specific endpoint URLs.}
\label{tab:agentic_sandbox_config}
\end{table*}

\section{Additional Analyses under Budget Pressure}
\label{app:budget_pressure}

This appendix reports four deterministic analyses and two trajectory diagnostics that support Section~\ref{sec:main_results}. Each statistic is
first computed within a domain. We then take an unweighted mean over the
domains that satisfy that metric's eligibility rules. We call this an
available-domain macro. 

\subsection{Oracle Portfolio}
\label{app:allocation_details}

The figure reports the difficulty composition of the curve-oracle
portfolio. Easy problems exceed their 50\% suite share in all 12 model--budget
portfolios. Hard problems remain below their \(1/6\) suite share in all 12
portfolios.

\begin{figure*}[!t]
    \centering
    \includegraphics[width=\textwidth]{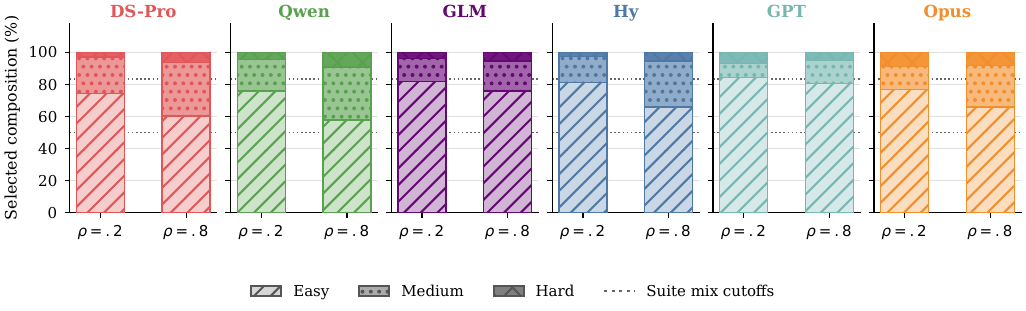}
    \caption{Difficulty composition of the curve-oracle portfolio. Dotted
    cutoffs mark the fixed suite mix: 50\% Easy, 33\% Medium, and 17\% Hard.
    This oracle-only diagnostic uses all three domains; missing domains are not
    imputed. Hy Math and AR oracle-dependent values inherit conservative
    lower-bound oracles.}
    \label{fig:oracle_difficulty}
\end{figure*}

\subsection{Selected Misses and Output Accrual by Position}
\label{app:cumulative_share}

\begin{figure*}[!t]
    \centering
    \includegraphics[width=\textwidth]{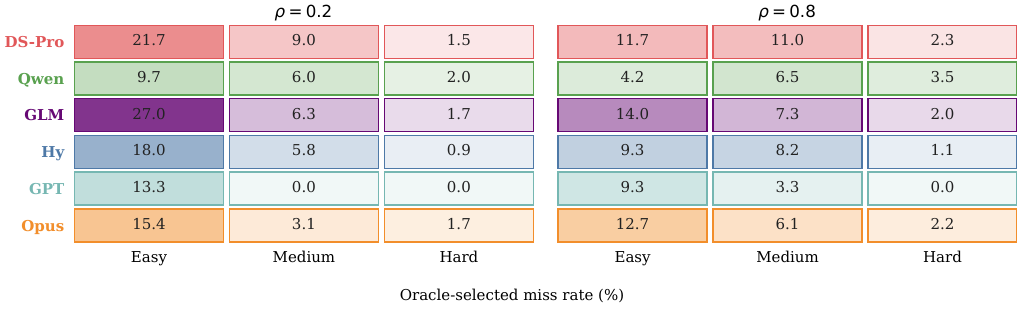}
    \caption{Curve-oracle selected-miss rates by budget pressure and difficulty tier.
    Each domain rate uses all judged problems in the explicitly paired suites
    as its denominator. Values are available-domain macros and missing domains
    are not imputed. Hy Math and AR oracle-dependent values inherit
    conservative lower-bound oracles.}
    \label{fig:selected_misses}
\end{figure*}

Figure~\ref{fig:selected_misses} reports
\emph{curve-oracle selected misses}. These are problems selected by the curve
oracle but missed in the contest. The rate divides the selected-miss count by
all judged problems in the same paired suites. At least one of the Easy or
Medium rates exceeds the Hard rate in all 12 model--budget panels; both do so
in 11. 

\begin{figure}[t]
    \centering
    \includegraphics[width=\columnwidth]{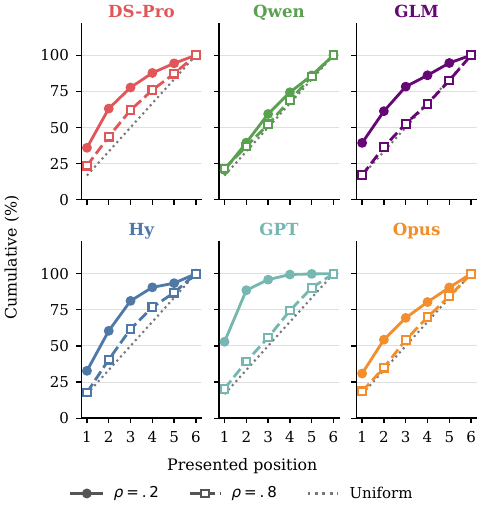}
    \caption{Cumulative attributed-output share by presented position for all six
    models at both budget pressures. The dotted line is a uniform
    \(1/6\)-per-position reference. Values are available-domain macros; missing
    domains are not imputed.}
    \label{fig:cumulative_share}
    \label{fig:cumulative_share_appendix}
\end{figure}

Figure~\ref{fig:cumulative_share} compares cumulative attributed-output share by position with a
uniform \(1/6\)-per-position reference. Attributed output is the output charged
to the budget, which for thinking models is the Stage~1 completion tokens and
therefore includes reasoning tokens. Under strong pressure, the first
position consumes 20.5--52.9\% of attributed output across the six available-
domain macros. Under moderate pressure, it consumes 16.9--23.5\%. Strong
pressure has the larger first-position share for five of the six models; Qwen
is the exception. 

\subsection{Wasted Output}
\label{app:wasted_output}

\begin{figure*}[!t]
    \centering
    \includegraphics[width=\textwidth]{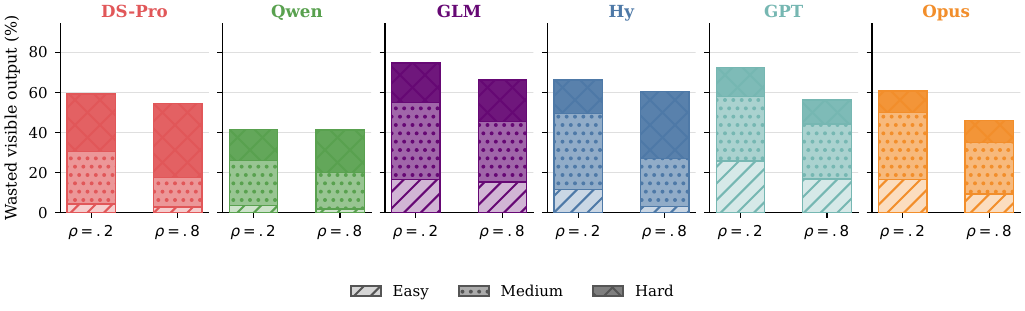}
    \caption{Six-model wasted attributed-output share at both budget pressures,
    stacked by difficulty. Wasted output is attributed output spent on ultimately
    incorrect problems. Values are available-domain macros. Missing domains are
    not imputed.}
    \label{fig:wasted_share}
\end{figure*}

Figure~\ref{fig:wasted_share} measures the attributed-output share spent on
problems that are ultimately incorrect. Total waste ranges from 41.2\% to
74.6\% across the 12 available-domain model--budget panels. The plotted total
is lower at \(\rho=.8\) than at \(\rho=.2\) for all six models. Its difficulty
composition still varies across models and budgets.

\subsection{Omission and Position Diagnostics}
\label{app:judge_audit}
\begin{figure*}[!t]
    \centering
    \includegraphics[width=\textwidth]{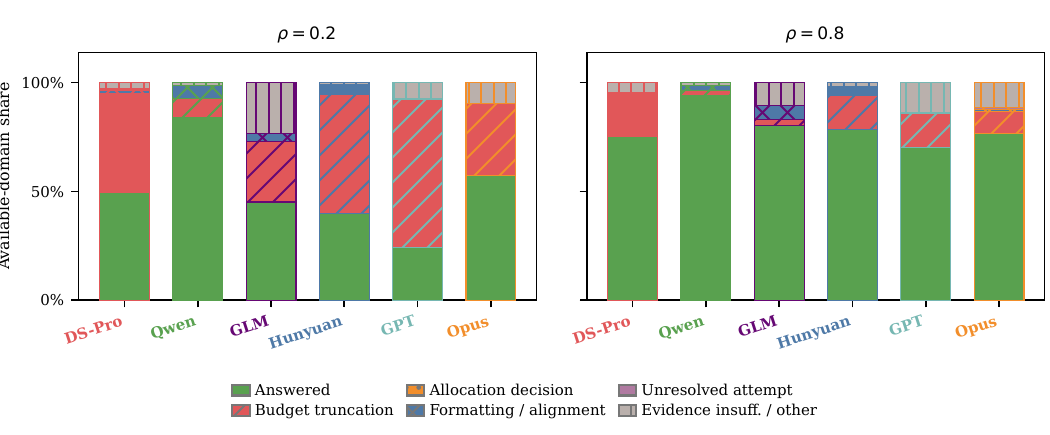}
    \caption{Available-domain decomposition of pure-NL problem slots at the
    two pressure levels. Bars separate answered/not-missing problems from
    budget truncation, formatting or alignment failures, unresolved attempts,
    and evidence-insufficient or other cases. Missing domains and trajectories
    are omitted rather than imputed. The allocation-decision category requires
    decision-time budget checkpoints and is therefore not identifiable here.}
    \label{fig:judge_omission}
\end{figure*}

Figure~\ref{fig:judge_omission} decomposes every available problem slot into
answered/not missing, budget truncation, formatting or alignment failure,
unresolved attempt, and residual evidence-insufficient cases. Under moderate
pressure, the answered share is higher and the truncation share is lower for
all six models. Under strong pressure, truncation ranges from 8.7\% to 68.0\%
across model macros; under moderate pressure it ranges from 2.5\% to 19.3\%.

\begin{figure*}[!t]
    \centering
    \includegraphics[width=\textwidth]{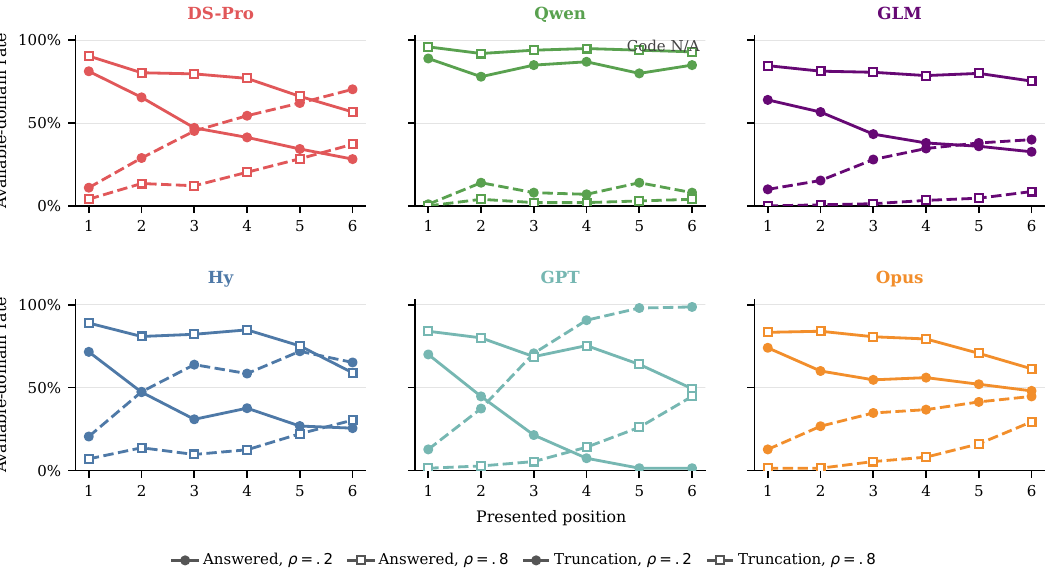}
    \caption{Pure-NL answered/not-missing and budget-truncation rates by
    presented position. Each panel is one model; solid lines show answered
    rates and dashed lines show truncation, with circles for \(\rho=.2\) and
    squares for \(\rho=.8\). Each point is an available-domain macro. Missing
    domains and trajectories are omitted rather than imputed. Answered rates
    fall and truncation rates rise from the first to the last position in all
    12 model--pressure series.}
    \label{fig:judge_position}
\end{figure*}

Figure~\ref{fig:judge_position} gives the trajectory-level counterpart to the
accuracy curves in Figure~\ref{fig:position_effect}. The answered/not-missing
rate falls from position 1 to position 6 in every model--pressure series, while
the judged budget-truncation rate rises in all 12 series. The change is usually
larger under strong pressure. 

\section{Human Annotation of Trajectories}
\label{app:trajectory-annotation}

\paragraph{Scope and annotation unit.}
Ten human annotators collectively examined every tool-free and agentic contest trajectory line by line under a shared codebook and produced all behavioral labels reported in this paper. The corpus covers six models, two budget pressures, and three domains, forming 36 model--pressure--domain cells. 

\paragraph{Corpus and annotation load.}
Each cell contains 50 contests with five independent trajectories per contest, both tool-free and agentic settings, so the annotated corpus is \(3\times 50\times 6\times 5\times 2\times 2 = 18000\) trajectories. Annotation is exhaustive rather than sampled. The corpus was partitioned evenly across the ten annotators, 1800 trajectories
each, by random assignment rather than by model or domain block. Model identity
was withheld: the standardized record carries no provider name, model name, or
interface metadata were removed from the standardized records, as were difficulty tiers and filesystem paths.

\paragraph{Single annotation.}
Each trajectory is annotated once. The codebook resolves every label into a
check on explicit trajectory evidence rather than a graded judgement: a
positive label requires an exact supporting quotation together with its step or
event identifier, and when a required component is absent or only speculative
the negative or insufficient-evidence label is mandatory. Verifier correctness,
reward, and oracle fields are frozen inputs that annotators cannot revise.

\paragraph{Annotation materials and procedure.}
For each trajectory, annotators inspected a standardized record containing the visible trajectory steps, action and budget events, submitted-artifact excerpts, verifier outcomes, oracle metadata, and exception information. Construction-time difficulty tiers and local filesystem paths were excluded. Verifier correctness, reward, and oracle fields were treated as frozen metadata: they established outcome facts and oracle applicability but could not substitute for behavioral evidence. Annotators did not re-solve problems or revise correctness judgments.

Annotations covered engineering issues, decision-regret patterns, online
adaptation, and oracle-gap attribution. Every positive behavioral label
required an exact supporting quotation and its trajectory step or event
identifier. When a required evidence component was absent or only speculative, annotators assigned the negative or insufficient-evidence label. Except for the primary oracle-gap category, labels were not required to be mutually exclusive. Engineering issues were restricted to API, runtime, scaffold, tool-parsing, container, trajectory-file, or verifier failures; ordinary reasoning errors and failed solution attempts were not engineering issues.

\paragraph{Oracle-gap attribution rubric.}
For each oracle-selected miss, annotators assigned exactly one primary cause:
\emph{never attempted}; \emph{attempted too late}; \emph{stopped after partial
progress}, where work began but no answer was submitted; \emph{spent budget
elsewhere}, where the shared budget was consumed on other problems;
\emph{misread tool feedback}; \emph{wrong format or incomplete finalization};
or \emph{genuinely unsolved}, where the trajectory shows no path to a correct
answer within the budget. Misses whose oracle metadata or behavioral evidence
was insufficient were recorded as such and excluded from the reported shares,
so the denominators are behavior-attributable misses rather than all misses.

\paragraph{Online-adaptation rubric.}
We distinguish four behavioral events:
\begin{itemize}
    \item \textbf{Difficulty observation} requires an explicit observation of
    task difficulty, uncertainty, a negative outcome, or tool/test feedback.
    Merely stating the remaining budget does not qualify.

    \item \textbf{Substantive strategy update} requires one focal episode with
    (i) a concrete strategy that has already been adopted or enacted,
    (ii) a genuinely new subsequent observation, and (iii) a substantive
    revision to the target, method, execution order, verification or
    finalization procedure, or resource allocation. The temporal requirement is
    \(t_{\mathrm{prior}} < t_{\mathrm{trigger}}
    \leq t_{\mathrm{revision}}\). Initial planning and bookkeeping-only changes
    to focus or shelving status do not qualify.

    \item \textbf{Resource-rational update} additionally requires an explicit
    causal link between the revision and newly observed feedback, expected
    value, cost, success probability, task difficulty, or the opportunity cost
    imposed by the remaining shared budget.

    \item \textbf{Attempted-but-failed update} requires the revised strategy to
    be enacted and the same targeted shortfall to remain visible in a strictly
    later observation. A fully correct run, a non-perfect aggregate score, or
    an engineering termination alone cannot establish this label.
\end{itemize}

\paragraph{Decision-regret rubric.}
We also annotate five descriptive counterfactual-regret patterns:
\begin{itemize}
    \item \textbf{Undercoverage} requires concrete uncovered work with
    non-trivial expected value, a feasible action at an identified decision
    point, and an allocation or abandonment decision that left the work
    uncovered.

    \item \textbf{Overcommitment} requires repeated disproportionate spending
    on one problem despite weak or negative feedback while alternative
    problems remained available.

    \item \textbf{Late budget exhaustion} requires both that exhaustion blocked
    intended work or finalization and that it was avoidable through earlier
    stopping, convergence, or consolidation.

    \item \textbf{Premature shelving} requires a shallow abandonment decision
    together with a concrete and feasible next action that was not pursued.

    \item \textbf{Weak finalization} requires submission-ready content for a
    specific problem and direct evidence that the corresponding submission was
    missing, malformed, or incomplete.
\end{itemize}
These labels characterize evidence visible in the trajectory; they do not identify causal effects.

\section{Behavioral Coordinate}
\label{app:behavioral-coordinate}

Figure~\ref{fig:coo} summarizes two complementary aspects of shared-budget
behavior for the six flagship models: how broadly a model works across the
suite and whether it revises its strategy in response to runtime evidence.
Each coordinate combines the tool-free and agentic settings, both budget
pressures, and all available domains. It is therefore not an agentic-only
statistic.

\paragraph{Problem coverage.}
For model \(m\), setting \(s\), pressure \(\rho\), and domain \(d\), let
\(\mathcal{T}_{m,s,\rho,d}\) denote the available clean contest records.
For each contest \(t\) and problem \(p\in\{1,\ldots,6\}\), the tri-state
variable \(X_{t,p}\in\{\mathrm{true},\mathrm{false},\mathrm{unknown}\}\)
records whether the trajectory contains visible, substantive
problem-specific work. Substantive evidence includes problem-specific
reasoning, a changed solution artifact, or an executed substantive action.
Focus and shelving commands, attribution bookkeeping, guard-only actions, and
final answers without supporting work do not establish coverage.

The confirmed and possible coverage rates within a cell are
\[
C^{-}_{m,s,\rho,d}
=
\frac{
\sum_{t\in\mathcal{T}_{m,s,\rho,d}}
\sum_{p=1}^{6}
\mathbf{1}[X_{t,p}=\mathrm{true}]
}{
6|\mathcal{T}_{m,s,\rho,d}|
},
\]
and
\[
C^{+}_{m,s,\rho,d}
=
\frac{
\sum_{t\in\mathcal{T}_{m,s,\rho,d}}
\sum_{p=1}^{6}
\mathbf{1}[X_{t,p}\neq\mathrm{false}]
}{
6|\mathcal{T}_{m,s,\rho,d}|
}.
\]
Figure~\ref{fig:coo} uses the confirmed lower bound \(C^{-}\).

\paragraph{Resource-rational strategy updates.}
For each clean contest \(t\), let
\(Y_t\in\{\mathrm{true},\mathrm{false},\mathrm{unknown}\}\) indicate whether
the trajectory contains a resource-rational strategy update. An initial plan
is not an update. A positive label requires a later observation to trigger a
substantive revision to the target problem, solution method, execution order,
attempt-or-stop policy, verification or finalization procedure, or resource
allocation. Bookkeeping-only changes do not qualify. The complete rubric is
given in Appendix~\ref{app:trajectory-annotation}.

The cell-level confirmed and possible rates are
\[
\begin{aligned}
R^{-}_{m,s,\rho,d}
&=
\frac{
\sum_{t\in\mathcal{T}_{m,s,\rho,d}}
\mathbf{1}[Y_t=\mathrm{true}]
}{
|\mathcal{T}_{m,s,\rho,d}|
},\\
R^{+}_{m,s,\rho,d}
&=
\frac{
\sum_{t\in\mathcal{T}_{m,s,\rho,d}}
\mathbf{1}[Y_t\neq\mathrm{false}]
}{
|\mathcal{T}_{m,s,\rho,d}|
}.
\end{aligned}
\]
This is the unconditional rate
\(P(\text{rational update})\) over all clean contests, not the conditional
rate \(P(\text{rational update}\mid\text{strategy update})\). Figure
~\ref{fig:coo} plots \(R^{-}\).

\paragraph{Quadrant interpretation.}
The dashed lines at \(C=0.5\) and \(R=0.5\) are descriptive references. They define four mnemonic regions:
\emph{Portfolio Planner} denotes broad coverage with frequent confirmed
resource-rational updates; \emph{Strategic Specialist} denotes narrower
coverage with frequent updates; \emph{Shallow Scanner} denotes broad coverage
but infrequent updates; and \emph{Tunnel Solver} denotes both narrow coverage
and infrequent updates. These names describe coordinate regions rather than
latent model types.

The main separation in Figure~\ref{fig:coo} is between coverage and adaptive
control. A model may perform substantive work on many problems while rarely
using new evidence to revise how it spends the remaining budget. Broad
coverage therefore does not by itself imply resource-rational reasoning.
\section{Online Scheduler Directives}
\label{app:online_scheduler_directives}

This appendix gives the implementation details for the online scheduler
directives used in Section~\ref{sec:recovering-allocation-gap}. The directives test whether simple runtime
control over cross-problem scheduling can recover part of the allocation gap
while holding fixed the underlying model, problem suite, budget, tool
interface, parser, and final judging protocol.

\paragraph{Compared strategies.}
We compare the contest reference with two scheduler directives in the same
budgeted agentic loop. The contest reference is the original baseline: the agent
is instructed to solve as many problems as possible and may choose the order
freely. Strategy~A adds a coverage guard, which prevents repeated paid investment
in one problem before every visible problem receives an initial focused probe.
Strategy~B keeps the coverage guard and adds a lightweight verification
gate. Once coverage is complete and a stable candidate answer or
solution file exists, this gate limits additional paid checking on the
unchanged candidate unless the candidate is rewritten. Strategies share the same coverage guard and differ only in the verification gate.
Strategies~A and~B do not use answer labels, hidden correctness
feedback, response-curve information, oracle-selected problems, or
difficulty labels. They operate only on runtime-visible state: the
currently focused problem, per-problem paid-step counts, the shared
budget counter, and the stability of the answer or solution artifact.
Thus, the intervention is an online scheduling constraint in the real
loop, not an oracle scheduler.

\paragraph{Coverage directive.}
The coverage directive is implemented as both an instruction and a runtime
guard. Its operative rule is: before spending a second paid focused step on the
same problem, give every visible problem at least one paid focused probe. The
runtime enforces this rule by tracking paid focused steps for each visible
problem. If the current problem has already received the allowed initial probe
and another visible problem has not, further paid actions on the current
problem are blocked until the agent moves to an unattempted problem. The
bookkeeping commands used for this attribution, such as
\texttt{focus\_problem} and \texttt{shelve\_problem}, are free and expose no
correctness signal.

\paragraph{Coverage plus verification directive.}
Strategy~B retains the coverage guard and adds a cheap verification gate. Once
all visible problems have received a paid probe, a problem with a stable
candidate answer or solution artifact may receive at most one additional paid
checking step unless the candidate changes. Rewriting the answer or solution
resets this allowance.

The gate is intentionally lightweight and non-oracle. It does not judge whether
a candidate is correct; it only limits repeated paid checking after the
candidate artifact has stabilized. In code, these checks may include local
compilation, execution, sample tests, or self-generated tests. In mathematics
and abstract reasoning, they may include numerical substitution, small-case
checks, enumeration, constraint checks, or format checks when such signals are
available. Hidden tests, official verdicts, proof oracles, and response-curve
information are never exposed during the run.

\paragraph{Budget accounting and enforcement.}
The online scheduler experiments use the same shared-budget agentic loop and
budget-accounting mechanism as the main benchmark. Strategies~A and~B add
scheduler guards on top of this loop, but do not change the tool whitelist,
artifact format, parser, judge, or final scoring protocol. Mathematics and
abstract reasoning use the \texttt{compute\_tools} policy for this intervention
study, while the code domain uses a problem-scoped counted-tool budget.
Detailed action-accounting rules are given in Appendix~\ref{app:white_list}.

\paragraph{Experimental protocol.}
The scheduler-directive study covers three models---DeepSeek-V4-Pro, GLM-5.2, and Hy-3---at strong budget pressure
\(\rho=0.2\). For each model, domain, and configuration, the evaluation contains
50 six-problem suites. The reported score is the mean number of correct answers per six-problem
suite over the 50 suites. Missing or unscored suites are counted as zero in the aggregate
mean. For code, the normalized reward is the number of solved problems
divided by the six problems in the suite; mathematics and abstract
reasoning use the same normalized six-problem scoring convention.

The reported values are computed from the contest-reference, A, and B runs for
each model under this protocol, using the same final judging and aggregation
procedure as the main agentic evaluation.

\section{Trajectory-Level Case Studies of Allocation Failure and Recovery}
\label{app:qualitative_cases}
This appendix provides trajectory-level illustrations of the aggregate
failure mechanisms reported in Sections~4 and~5. The examples show how allocation failures unfold during a shared-budget episode; prevalence is estimated separately from the full-corpus analyses in the main text. We manually audited 14 shortlisted candidate cases identified by the aggregate diagnostic pipeline and selected examples with formal outcomes, complete trajectories and artifacts, and no detected infrastructure failure. Selection criteria were evidentiary completeness and mechanistic distinctness.

Final correctness is always taken from the domain-specific parser or verifier.
Behavioral labels are used only for retrospective diagnosis. Response-curve
costs and oracle selections were not visible to the model. Accordingly, oracle
evidence indicates observed allocation headroom, not a guarantee that the
oracle portfolio could be reproduced in one contest trajectory.
Appendix~\ref{app:oracle_knapsack} defines the replay and its limitations.

\subsection{Case I: Local Debugging Crowds Out Four Oracle-Selected Problems}
\label{app:case_overcommitment}

Our first case illustrates how repeated local repair can produce both
overcommitment and undercoverage. DeepSeek-Chat runs in the agentic coding
setting on Code Set~47 at $\rho=0.2$. The shared budget is ten counted
actions. The agent spends all ten actions on Problem~C, obtains no accepted
solution, and leaves the other five problems untouched. The final contest
score is $0/6$. In contrast, the cap-limited response-curve oracle has value
$4/6$: it selects four problems with a total observed cost of eight actions.

\begin{table*}[!t]
\centering
\scriptsize
\setlength{\tabcolsep}{2.5pt}
\renewcommand{\arraystretch}{1.08}
\begin{tabular}{
p{0.10\textwidth}
p{0.07\textwidth}
p{0.08\textwidth}
p{0.14\textwidth}
p{0.07\textwidth}
p{0.08\textwidth}
p{0.17\textwidth}
p{0.16\textwidth}
}
\toprule
Problem & Tier & Oracle? & Lowest observed successful cost &
Contest actions & Outcome & State & Diagnosis \\
\midrule

A: 2120E
& Medium
& Yes
& 2
& 0
& Missing
& Never focused
& Cheap oracle-selected miss \\

B: 2091B
& Easy
& Yes
& 2
& 0
& Missing
& Never focused
& Cheap oracle-selected miss \\

C: 2107A
& Easy
& No
& 3
& 10
& Wrong
& Repeated construction, repair, compilation, and testing
& Overcommitment target \\

D: 2029B
& Easy
& Yes
& 2
& 0
& Missing
& Never focused
& Cheap oracle-selected miss \\

E: 1992F
& Medium
& Yes
& 2
& 0
& Missing
& Never focused
& Cheap oracle-selected miss \\

F: 2107F2
& Hard
& No
& No observed success
& 0
& Missing
& Never focused
& No response-curve reachability evidence \\

\bottomrule
\end{tabular}
\caption{Problem-level allocation in the overcommitment case. Costs are
counted actions. The lowest observed successful cost is the cheapest grid level
of the sampled single-problem response curve at which the problem was solved in
at least one repeat, not the unknown theoretical minimum required to solve the
problem.}
\label{tab:qual_case_overcommitment}
\end{table*}

The trajectory begins with a free status query and an explicit
acknowledgement of the budget: \emph{``I have 10 counted steps. Let me start
with Problem C.''} After focusing Problem~C, the agent never switches or
shelves it. Actions 1--5 are spent constructing and rewriting the program.
Actions 6--8 repair file-construction and compilation failures. Action~9
finally produces a compilable candidate, and Action~10 runs a local sample
test. The full budget is therefore exhausted on one incorrect solution while
the other five problems remain untouched.

The decisive issue is not simply that one problem receives many actions.
Problem~C ultimately produces zero reward, while Problems~A, B, D, and E each
have an accepted single-problem response-curve point at an observed cost of two
actions. Under the minimum-cost tie-break used by our implementation, the
saved selected optimum is unique: Problems~A, B, D, and E, with a total
observed cost of eight actions. Relative to this replay, continued local repair
on C displaces four empirically reachable opportunities. Notably,
overcommitment does not require a nominally hard target: Problem~C belongs to
an easy benchmark tier. The agent responds to each local error but never
reassesses whether another problem has become a better use of the remaining
budget.

\subsection{Case II: More Attributed Output than a Successful Standalone Run}
\label{app:case_transfer_failure}

The second case illustrates a single-to-contest transfer failure in tool-free coding. DeepSeek-Chat is evaluated on Code Set~9 at
$\rho=0.8$. The model obtains two accepted solutions in the contest, while
the response-curve oracle selects three. The missed oracle-selected item is
Problem~F, Codeforces 2000A.

The same model solves Problem~F in a standalone response-curve run using
173 output tokens. In the contest, proportional problem-section attribution
assigns 322.4 Stage-1 tokens to the same problem, yet the resulting program
receives a wrong-answer verdict. The audit pipeline also records a broader
446.3-token proxy that includes Stage-2 formatting output; all mechanism
claims below use the Stage-1-only value.

Stage~1 terminates normally after generating 1,618 of the available 5,065
tokens, so the failure is not caused by token-cap truncation. Stage~2 performs
extraction only and reproduces the same incorrect code. The standalone run
checks the required prefix and validates the exponent suffix without requiring
the suffix to consist entirely of zeros. By contrast, the contest run
interprets the notation as requiring every character after the initial
\texttt{10} to be \texttt{0} and implements this stronger rule.

The contest reasoning states that the string must begin with \texttt{10} and
that \emph{``all digits after the first two must be `0'.''} The corresponding
loop rejects a candidate whenever any later character is nonzero. This is a
task misinterpretation rather than an incomplete implementation. The formal
verifier marks the complete contest program wrong, whereas the standalone
program is accepted.

The model thus receives more attributed Stage-1 tokens than its observed
successful standalone cost, emits a complete program, and is not truncated,
yet selects a different, incorrect interpretation. The broader pipeline proxy
leads to the same conclusion, but we use Stage~1 because Stage~2 only extracts
the candidate. The 173-token value is the lowest sampled successful point, not
a theoretical minimum, and the 322.4-token value is a section-length
attribution rather than an API counter. The evidence supports a funded
transfer failure, not a claim that token quantity determines whether the
standalone reasoning path is reproduced.

\subsection{Case III: Protocol-Shaped Shelving Followed by Weak Finalization}
\label{app:case_weak_finalization}

The third case comes from an abstract-reasoning run under a coverage-first
runtime policy that uses the same initial-coverage guard studied in
Appendix~\ref{app:online_scheduler_directives}. We use it primarily to
illustrate weak finalization under a coverage-first policy rather than an
unconfounded baseline shelving failure. DeepSeek-V4-Pro (DS-Pro) is evaluated
on AR Contest~9 at $\rho=0.8$, with a budget of seven paid compute actions.
The agent uses six actions, answers three of six problems correctly, and
terminates with one action unused. This run is not one of the fixed
$\rho=0.2$ cells reported in the scheduler intervention study.

The relevant item is Problem~3, a palindrome-generation task. During its first
paid probe, the agent constructs a string that is itself a palindrome but does
not use all required input letters. The agent explicitly records this defect
when shelving the problem:

\begin{quote}
\small
\texttt{Palindrome: [...] -- but doesn't use all letters due to multiple
odd-count chars.}
\end{quote}

A deterministic post-hoc multiset check confirms the defect. The input
contains 68 letters, whereas the candidate contains 65 and omits one occurrence
each of \texttt{h}, \texttt{k}, and \texttt{r}. This check was not exposed to
the model during the run.

\begin{table}[!htbp]
\centering
\scriptsize
\setlength{\tabcolsep}{3pt}
\renewcommand{\arraystretch}{1.08}
\begin{tabular}{
p{0.18\columnwidth}
p{0.46\columnwidth}
p{0.24\columnwidth}
}
\toprule
Stage & Evidence and decision & Budget and consequence \\
\midrule

Initial probe
& The candidate is a palindrome, but the all-letters constraint is known to
be violated; the agent attempts further work on Problem~3
& 4 remain; progress is substantive but incomplete \\

Coverage guard
& A second paid action is requested before full initial coverage; the runtime
blocks it and requires a switch
& 4 remain; the first switch is protocol-shaped \\

Shelving
& The agent records the defect and continues the coverage pass
& 4 remain; the candidate is stored but unresolved \\

Coverage completion
& Problems~4--6 each receive one paid probe; return to Problem~3 is now
permitted
& 1 remains \\

Finalization
& The agent notes that one compute step remains and considers revisiting
Problem~3, but keeps the current answer
& 1 unused; the candidate receives non-full credit and is counted incorrect \\

\bottomrule
\end{tabular}
\caption{Progress, protocol-shaped shelving, and finalization decisions in the
abstract-reasoning case.}
\label{tab:qual_case_finalization}
\end{table}

The two decisions require different interpretations. The initial attempt to
reinvest in Problem~3 is blocked because the remaining problems have not
received initial probes, so the first switch is protocol-shaped rather than an
unconfounded voluntary give-up. After all six problems are covered, however,
the guard no longer prevents a return and one paid compute action remains.
The agent considers investigating the known defect but states,
\emph{``I'll keep my current answer.''} The same invalid candidate reaches the
final artifact and receives a formal score of $0.05$, which is counted as
incorrect under the full-credit binarization used in the main results. We
therefore interpret the case as protocol-shaped shelving followed by
model-chosen weak finalization. Because the run has no linked per-contest
response-curve oracle record, it illustrates failure to improve known partial
progress rather than a numerical oracle-recoverable gap.

\subsection{Matched Intervention Example: Broader Coverage with Two Additional
Accepted Answers}
\label{app:case_scheduler_matched}

Finally, we provide a matched example from the online-scheduler intervention.
The comparison uses DS-Pro on the same Code Set~12, with the same fixed budget
of seven counted actions, tools, parser, and verifier. This is the budget
\(A^{0.2}_{\text{DS-Pro},\mathrm{Code}}\) of the main benchmark, so the case runs
under the reported configuration rather than a separate one. The contest
reference is the baseline agent. Strategy~B adds the coverage guard and the
verification gate defined in
Appendix~\ref{app:online_scheduler_directives}.
\begin{table}[t]
\centering
\scriptsize
\setlength{\tabcolsep}{3pt}
\renewcommand{\arraystretch}{1.08}
\begin{tabular}{
p{0.18\columnwidth}
p{0.18\columnwidth}
p{0.18\columnwidth}
p{0.35\columnwidth}
}
\toprule
Problem & Contest reference & Strategy~B & Allocation effect \\
\midrule

A: 2108B (M)
& 3 / Wrong
& 2 / Wrong
& One fewer action on the initial target; repeated low-return investment is
curtailed \\

B: 2070B (E)
& 3 / Accepted
& 1 / Accepted
& Existing success is preserved with two fewer actions \\

C: 2003D2 (H)
& 0 / Missing
& 1 / Wrong
& Newly covered, but coverage does not guarantee success \\

D: 2070C (E)
& 0 / Missing
& 1 / Accepted
& Newly covered; one answer is recovered \\

E: 2022E1 (M)
& 0 / Missing
& 1 / Accepted
& Newly covered; a second answer is recovered \\

F: 2000C (E)
& 1 / Accepted
& 1 / Accepted
& Existing success is preserved at unchanged cost \\

\bottomrule
\end{tabular}
\caption{Matched scheduler comparison on the same coding contest. Both
strategies execute seven counted actions. The contest reference scores $2/6$,
while Strategy~B scores $4/6$. Parenthetical letters denote benchmark tiers.}
\label{tab:qual_case_scheduler}
\end{table}

The contest reference allocates its actions as $(3,3,0,0,0,1)$ across
Problems A--F. It concentrates six of its seven actions on Problems~A and~B and
never touches Problems~C--E. Strategy~B allocates $(2,1,1,1,1,1)$; its coverage
guard blocks three attempted reinvestments before execution, and blocked
commands consume no budget. Problems~D and E, missing under the contest
reference, are accepted under Strategy~B.

The pair illustrates the mechanism behind the aggregate coding improvement:
limiting early concentration exposes additional solvable problems. The budget
makes the trade-off sharp. Because the guard requires a paid probe on each
problem, covering all six consumes six of the seven available actions, so
Strategy~B retains a single action for additional depth and spends it on
Problem~A. It also shows the cost of breadth, since one action goes to the
unsolved hard Problem~C and Problem~B keeps its accepted answer with two fewer
actions than before. The trajectories are
independent model samples, so the comparison is not a deterministic
counterfactual. The mechanism also does not transfer uniformly across domains:
in aggregate the directives improve Code for every model, whereas in mathematics
and abstract reasoning they help some models and hurt others
(Table~\ref{tab:online_scheduler_directives}). Moreover, the observed change is
driven primarily by the coverage constraint and does not isolate the incremental
effect of the verification gate.

\section{Target-in-Suite Context-Stress Diagnostic}
\label{app:context-stress}

A possible alternative explanation for the contest--oracle gap is that part of the deficit arises from suite-context interference caused by presenting six problem statements, rather than from cross-problem resource allocation. We therefore conduct a target-in-suite diagnostic that preserves the full six-problem context while removing the need to select among or allocate resources across multiple actionable problems.

\paragraph{Experimental design.}
We evaluate Qwen3.7-Max and DeepSeek-V4-Pro in both tool-free and agentic settings across mathematics, competitive programming, and abstract reasoning. For each domain, we select 10 suites by deterministic length-stratified sampling based only on public statement length and a frozen seed; the same suites are shared across models and settings. Each of the six positions serves once as the target, yielding 60 target episodes per model--setting--domain cell and 720 stress episodes in total. The model sees all six problem statements in their original order, but only one is designated for solution and scored, while the other five are marked as distractors. In the agentic setting, computation on distractor problems is additionally blocked. Each target receives the largest frozen single-problem budget in our response-curve evaluation.

\paragraph{Controls and analysis.}
Each stress outcome is paired with a problem-matched single-problem reference from the corresponding response-curve evaluation. We first average the six paired control--stress differences within each suite and then average equally over the 10 suites. We treat this experiment as a robustness diagnostic rather than a causal decomposition of the contest--oracle gap.

\paragraph{Results.}
Table~\ref{tab:context-stress} reports the target-in-suite diagnostic results.

\begin{table}[!t]
\centering
\footnotesize
\setlength{\tabcolsep}{3pt}
\begin{tabular}{@{}lcrrr@{}}
\toprule
Model & Domain & Control & Stress & $\Delta$ (pp) \\
\midrule
\multicolumn{5}{@{}l}{\textit{Agentic}} \\
\multirow{3}{*}{DeepSeek-V4-Pro}
& AR   & 0.800 & 0.857 & -5.7 \\
& Code & 0.917 & 0.867 &  5.0 \\
& Math & 0.867 & 0.867 &  0.0 \\
\cmidrule(lr){1-5}
\multirow{3}{*}{Qwen3.7-Max}
& AR   & 0.835 & 0.851 & -1.6 \\
& Code & 0.900 & 0.817 &  8.3 \\
& Math & 0.817 & 0.817 &  0.0 \\
\midrule
\multicolumn{5}{@{}l}{\textit{Tool-Free}} \\
\multirow{3}{*}{DeepSeek-V4-Pro}
& AR   & 0.700 & 0.742 & -4.1 \\
& Code & 0.867 & 0.883 & -1.7 \\
& Math & 0.850 & 0.850 &  0.0 \\
\cmidrule(lr){1-5}
\multirow{3}{*}{Qwen3.7-Max}
& AR   & 0.548 & 0.497 &  5.0 \\
& Code & 0.900 & 0.833 &  6.7 \\
& Math & 0.850 & 0.833 &  1.7 \\
\bottomrule
\end{tabular}
\caption{Target-in-suite context-stress results. Control and Stress are mean target scores; $\Delta=\text{Control}-\text{Stress}$ in percentage points, so positive values indicate lower performance under context stress.}
\label{tab:context-stress}
\end{table}

Across the 12 model--setting--domain cells, nine have point-estimated control--stress differences of at most 5 percentage points, with an unweighted descriptive mean difference of 1.1 percentage points. All six DeepSeek-V4-Pro cells have differences of at most 5 points. Qwen3.7-Max is more heterogeneous, with differences of 6.7 points in tool-free Coding and 8.3 points in Agentic Coding.

Overall, additional suite context does not consistently reduce target-only performance under explicit target cues and high single-problem budgets, arguing against a uniform suite-context explanation of the contest deficit.

\FloatBarrier

\end{document}